\documentclass[journal]{IEEEtran}
\usepackage{algorithm}
\usepackage[english]{babel}
\usepackage{blindtext}
\usepackage{algorithmic}
\usepackage{color, soul}

\usepackage{multirow}
\usepackage{xcolor}
\usepackage{amsfonts}
\usepackage{overpic}
\usepackage{epsfig}
\usepackage{graphicx}
\usepackage{amsmath}
\usepackage{amsmath,bm}
\usepackage{amssymb,subfigure,cases}
\usepackage{color,hyperref}
\usepackage{caption}

\ifCLASSINFOpdf
\else
\fi
\definecolor{darkblue}{rgb}{0.0,0.0,1.0}
\hypersetup{colorlinks,breaklinks,
            linkcolor=darkblue,urlcolor=darkblue,
            anchorcolor=darkblue,citecolor=darkblue}

\begin{document}

\title{Exploring Models and Data for\\
 Remote Sensing Image Caption Generation}

\author{Xiaoqiang~Lu,~\IEEEmembership{Senior Member,~IEEE,}
        Binqiang~Wang,
        Xiangtao~Zheng,\IEEEmembership{}
        and~Xuelong~Li,~\IEEEmembership{Fellow,~IEEE}
\thanks{This work was  supported  in  part  by the National Natural Science Foundation of China under Grant 61761130079, in  part by the Key Research Program of Frontier Sciences, CAS under Grant QYZDY-SSW-JSC044, in part by the National Natural Science Foundation of China under Grant 61472413, in part by the  National Natural Science Foundation of China under Grant 61772510, and in part by the Young Top-notch Talent Program of Chinese Academy of Sciences under Grant QYZDB-SSW-JSC015.

The authors are with the Xi'an Institute of Optics and Precision Mechanics,
Chinese Academy of Sciences, Xi'an 710119, Shaanxi, P. R. China and with
The University of Chinese Academy of Sciences, Beijing 100049, P. R. China.}

}

\maketitle

\begin{abstract}
Inspired by recent development of artificial satellite, remote sensing images have attracted extensive attention. Recently, noticeable progress has been made in scene classification and target detection.
However, it is still not clear how to describe the remote sensing image content with accurate and concise sentences. In this paper, we investigate to describe the remote sensing images with accurate and flexible sentences.
First, some annotated instructions are presented to better describe the remote sensing images considering the special characteristics of remote sensing images.
Second, in order to exhaustively exploit the contents of remote sensing images, a large-scale aerial image dataset is constructed for remote sensing image caption.
Finally, a comprehensive review is presented on the proposed dataset to fully advance the task of remote sensing caption. Extensive experiments on the proposed dataset demonstrate that the content of the remote sensing image can be completely described by generating language descriptions. The dataset is available at \emph{https://github.com/201528014227051/RSICD\_optimal}
\end{abstract}

\IEEEpeerreviewmaketitle

\section{Introduction}
\label{sec:1}
\IEEEPARstart{W}{ith} the development of remote sensing technology, remote sensing images have become available and attracted extensive attention in numerous applications \cite{Yanfeng2016Nonlinear, Gu2017Multiple, Xia2016, Yuan2015Substance, Lu2017Remote, lu201503, lu201601, lu201701, lu201702, cheng2016learning, yao2016semantic, han2015object, zheng2016target, yuan2017discovering, lu2017joint}. However, the studies of the remote sensing images are still concentrate on scene classification \cite{Lu2017Remote, cheng2015effective, tuia2009active, Yanfeng2015A, melgani2004classification}, object recognition \cite{penatti2015deep, inglada2007automatic} and segmentation \cite{yuan2014remote, meinel2004comparison, fan2009single, farag2005unified}. These works only recognize the objects in the images or get the class labels of the images, but ignore the attributes of the objects and the relation between each object \cite{Qu2016124}. To present semantic information of the remote sensing image, remote sensing image captioning aims to generate comprehensive sentences that summarize the image content in a semantic level \cite{Qu2016124}. This concise description of the remote sensing scene plays a vital role in numerous fields, such as image retrieval \cite{lu2017latent}, scene classification \cite{lu201702}, and military intelligence generation \cite{shi2017can}.

Image captioning is a difficult but fundamental problem in artificial intelligence. In the past few decades, many methods have been designed for natural image captioning \cite{Karpathy20153128, Xu20152048, Johnson20164565, ordonez2011im2text, Li2012, Yang2011444}. For natural image captioning, image representation and sentence generation are two aspects that the conventional methods is concentrated on. For image representation, deep convolutional representation has conquered the handcrafted representation. As for sentence generation, the studies has developed from traditional retrieved-based method to \emph{Recurrent Neural Network} (RNN).

To better describe the image content, many image representations, including static global representations and dynamic region representations, are considered. The global representations compress the whole image into a static representation \cite{Karpathy20153128}, while the dynamic region representations dynamically analyze the image content based on multiple visual regions \cite{ordonez2011im2text}. To decode the image representations into natural language sentences, several methods have been proposed for generating image descriptions \cite{Karpathy20153128, ordonez2011im2text, Li2012, Yang2011444}, such as \emph{Recurrent Neural Network} (RNN), \emph{Long-Short Term Memory networks} (LSTM), retrieve based method and object detection based method. All these methods have shown promising potentials in describing the images with concise sentences.

Although the aforementioned methods have achieved success in natural image captioning \cite{Hodosh2013853,Young201467,Chen2015, ordonez2011im2text}, they may be inappropriate for the remote sensing image captioning. Specially, the remote sensing image captioning is more complex than the natural image captioning \cite{Qu2016124, shi2017can}, and the semantics in remote sensing image become much ambiguous from the ``view of God''. For example, the remote sensing images are captured from airplanes or satellites, making the image contents complicated for ordinary people to describe.

Recently, some researchers have studied the remote sensing image caption and generated the sentences from the remote sensing image \cite{Qu2016124, shi2017can}. Qu \emph{et al.}  \cite{Qu2016124} firstly proposed a deep multimodal neural network model for semantic understanding of the high resolution remote sensing images.  Shi \emph{et al.} \cite{shi2017can} proposed a remote sensing image captioning framework by leveraging the techniques of \emph{Convolutional Neural Network} (CNN). Both methods used CNN to represent the image and generated the corresponding sentences from the trained model ( recurrent neural networks in \cite{Qu2016124} and  pre-defined templates in \cite{shi2017can}).
However, the generated sentences in both methods are simple, which cannot describe the complex content in remote sensing images as detailed as possible. Furthermore, the evaluations of remote sensing captioning are typically conducted on small datasets with a low diversity. The limited datasets are insufficient to approximate the remote sensing applications.

In this paper, we investigate to describe the remote sensing images with accurate and flexible sentences. \textbf{Firstly}, some special characteristics should be considered while annotating the sentences in remote sensing captioning:
1) Scale Ambiguity. The ground objects in remote sensing images may present different semantics under different scales.
2) Category Ambiguity. There are many categories in some regions of remote sensing images. It is hard to describe the fused regions with a single category label.
3) Rotation Ambiguity. The remote sensing image can be viewed from different rotations, since it is took from the ``view of God'' without a fixed direction.

\textbf{Then}, a large-scale aerial image dataset is constructed for remote sensing caption. In this dataset, more than ten thousands remote sensing images are collected from Google Earth \cite{Xia2016}, Baidu Map, MapABC, Tianditu. The images are fixed to 224$\times$224 pixels with various resolutions. The total number of remote sensing images are 10921, with five sentences descriptions per image. To the best of our knowledge, this dataset is the largest dataset for remote sensing captioning. The sample images in the dataset are with high intra-class diversity and low inter-class dissimilarity. Thus, this dataset provides the researchers a data resource to advance the task of remote sensing captioning.

\textbf{Finally}, a comprehensive review is presented on the proposed benchmark dataset (named as RSICD). Our dataset is publicly available in BaiduPan\footnote{http://pan.baidu.com/s/1bp71tE3}, Github\footnote{https://github.com/201528014227051/RSICD\_optimal} and Google Drive\footnote{https://drive.google.com/open?id=0B1jt7lJDEXy3aE90cG9YSl9ScUk} . In this paper, we focus on the encoder-decoder frameworks which are analogous to translating an image to a sentence \cite{Xu20152048}. To fully advance the task of remote sensing captioning, the representative encoder-decoder frameworks \cite{Xu20152048} are evaluated with various experimental protocols on the new dataset.

In summary, the main contributions of this paper contains three aspects:
\begin{itemize}
\item To better describe the remote sensing images, some special characteristics are considered: scale ambiguity, category ambiguity and rotation ambiguity.
\item A large-scale benchmark dataset of remote sensing images is presented to advance the task of remote sensing image captioning.
\item We present a comprehensive review of popular caption methods on our dataset, and evaluate various image representations and sentence generations methods using handcrafted features and deep feature.
\end{itemize}

The rest of this paper is organized as follows: In Section \ref{sec:2}, we review the related work of image captioning. The details of RSICD dataset are described in Section \ref{sec:3}. Then, we provide the description of encoder-decoder methods for benchmark evaluation in Section \ref{sec:4}. In Section \ref{sec:5}, the evaluation of baseline methods under different experimental settings are given. Finally, we summarize the conclusions of the paper in Section \ref{sec:6}.

\section{Related Work}
\label{sec:2}
\par
This section comprehensively reviews the existing image captioning including natural image captioning and remote sensing image captioning. Then the differences between natural image and remote sensing image are discussed. The difficulty of describing a remote sensing image and remote sensing image caption problem are presented at last.

\subsection{Natural image captioning}
Natural image captioning, which generates a sentence to describe a natural image, has been studied in computer vision for several years \cite{Karpathy20153128, Xu20152048, Johnson20164565, ordonez2011im2text, Li2012, Yang2011444}. The methods of the natural image captioning can be divided to three categories, including retrieved-based method, object detection based method and encoder-decoder method.

The retrieved-based method generated sentences based on the retrieval result \cite{ordonez2011im2text}. These methods search for similar images with the query image firstly, and then generate sentences based on the sentences which describe same kind images in datasets. In this case, the grammatical structure of the generated sentences is very poor, sometimes even unreadable. This is because the quality of generated sentences is closely dependent on the result of image retrieval. The retrieved result would be embarrassed when the content of the image is very different from the pictures in the database which are retrieved.

The second kind of method is based on object detection \cite{Li2012, Yang2011444}. Object detection based methods detect the objects in an image and then model the relation between the detected objects. The sentence is generated lastly by a sentence generating model using the information describing the relations of objects in image. The performance of the relation model is vital to the generated sentence. And the result of the object detection also has tremendous influence on the finally generated sentences.

The encoder-decoder methods \cite{Karpathy20153128, Xu20152048, Johnson20164565} complete the task in an end-to-end manner. These methods encoded an image to a vector firstly, and then utilized a model to decode the vector to a sentence. The deep neural models, usually using \emph{Convolutional Neural Network} (CNN) to extract features of images and then using language generating models such as \emph{Recurrent Neural Network} (RNN) and a special RNN called \emph{Long-Short Term Memory networks} (LSTM), have made a great progress in natural image captioning.

The encoder-decoder methods can generate promising sentences to describe the natural images. Inspired by recent advances in computer vision and machine translation \cite{Cho2014}, a deep model \cite{Vinyals20153156} is proposed to maximize the likelihood of the target description sentence given the training image. Karpathy \emph{et al.} \cite{Karpathy20153128} presented a \emph{Multimodal Recurrent Neural Network} architecture to generate concise descriptions of image regions. The \emph{Multimodal Recurrent Neural Network} architecture is a novel combination of three networks, including convolutional neural networks to extract image feature, bidirectional recurrent neural networks to represent sentences, and a structured objective embedding the image feature and sentence representation. Xu \emph{et al.} \cite{Xu20152048} proposed a caption generation method using two kinds of attention mechanisms to decide ``where" or ``what" to look in an image. Johnson \emph{et al.} \cite{Johnson20164565} introduced a caption task, which asked computer vision system to localize and describe outburst regions of a image using natural language. To further complete the task, Johnson \emph{et al.} \cite{Johnson20164565} proposed a \emph{Fully Convolutional Localization Network} (FCLN) architecture that can localize and describe regions of an image at the same time. Yang \emph{et al.} \cite{Yang20162361} proposed a novel encoder-decoder framework, called review network. It mimics a machine translation system, which translate the source language sentence into the target language sentence, take place the former sentence with image. Meanwhile, the review network also takes attention mechanism into consideration, by introducing attentive input reviewer and attentive output reviewer.

\subsection{Remote sensing image captioning}

Although many methods have been proposed for natural image captioning, only few studies on remote sensing image captioning can be focused \cite{shi2017can}. This is because there is not an accredited dataset like \emph{Common Objects in Context} (COCO) dataset in natural image datasets. Qu \emph{et al.}  \cite{Qu2016124} firstly proposed a deep multimodal neural network model to generate sentences for the high resolution remote sensing images. Shi \emph{et al.} \cite{shi2017can} proposed a remote sensing image captioning framework by leveraging the techniques of deep learning and \emph{Convolutional Neural Network} (CNN). However, the generated sentences of both methods are simple, which cannot describe the special characteristics in remote sensing images as detailed as possible. The special characteristics of remote sensing images are as follows:
\begin{itemize}
\item The regions in remote sensing scene may cover many land cover categories. For example, green plants are the main things we can see in remote sensing image. And the green plants include green grass, green crops, and green trees, which are difficult to distinguish one from another.
\item The objects in remote sensing images are different from those of natural images. Because there is no focus like that of natural pictures, we need to describe all the important things in a remote sensing image.
\item The objects in remote sensing exhibit rotation and translation variations. For a natural image, a building is usually from bottom to up and a person often stand on ground with their feet, hardly can we see a person with his head on the ground and his feet up to the sky. For remote sensing images, while, there is no difference between up and down, left and right. This is because it is hard for us to get the information how a remote sensing image be captured in most conditions. In this case, the information of the direction generated from a given remote sensing image is not clear.
\end{itemize}
Considered the above characteristics, a large-scale remote sensing captioning dataset is presented including more than ten thousands remote sensing images.

\section{Dataset for remote sensing image captioning}
\label{sec:3}
\par
In this section, we first reviews the existed remote sensing image captioning dataset, and then described the proposed \emph{Remote Sensing Image Captioning Dataset} (RSICD).

\subsection{Existing datasets for remote sensing image captioning}
\subsubsection{UCM-captions dataset}

This dataset is proposed in \cite{Qu2016124}, which is based on the UC Merced Land Use Dataset \cite{Yang2010270}. It contains 21 classes land use image, including agricultural, airplane, baseball diamond, beach, buildings, chaparral, dense residential, forest, freeway, golf course, harbor, intersection, medium residential, mobile home park, overpass, parking lot, river, runway, sparse residential, storage tanks and tennis court, with 100 images for each class. Every image is 256$\times$256 pixels. The pixel resolution of these images is 0.3048m. The images in UC Merced Land Use Dataset were manually extracted from many large images from the \emph{United States Geological Survey} (USGS) National Map Urban Area Imagery. Based on \cite{Qu2016124}, five different sentences were exploited to describe every image. The diversity of five coherent sentences for one image are totally different, but the sentence difference between images of the same class is very small.

\subsubsection{Sydney-captions dataset}

This dataset is also provided by \cite{Qu2016124}, which is based on the Sydney Data Set \cite{Zhang20152175}. The 18000$\times$14000 pixels large image of Sydney, Australia, was got from Google Earth. The pixel resolution of the image is 0.5m. It contains seven classes after cropping and selecting, including residential, airport, meadow, rivers, ocean, industrial, and runway. Similar with UCM-captions dataset, five different sentences were given to describe every image \cite{Qu2016124}. To describe a remote sensing image exhaustively, we should pay attention to the different attention of different people to an image and different sentence patterns. Both datasets, including UCM-captions and Sydney-captions, only focus on the latter problem, but the first problem should also be considered.

\subsubsection{A untitled dataset}

In \cite{shi2017can}, a untitled and undisclosed dataset about remote sensing image captioning is proposed. However, the sentences in \cite{shi2017can} are more like a fixed semantic modal added on multi-objects detection. The sentences of this dataset are just like ``this image shows an urban area", ``this image consists of some land structures", ``there is one large airplane in this picture" and so on. These sentences are lack of flexibility and diversity.

\subsection{RSICD: A new Dataset for Remote Sensing Image Captioning}
To advance the state-of-the-art performances in remote sensing image captioning, we construct a new remote sensing image captioning dataset, named RSICD, for remote sensing image captioning task. According to the variable scales and rotation invariant of remote sensing images, some instructions are provided to annotate RSICD dataset with comprehensive sentences.
\begin{figure*}[!t]
    \centering
    \scalebox{0.78}{\includegraphics{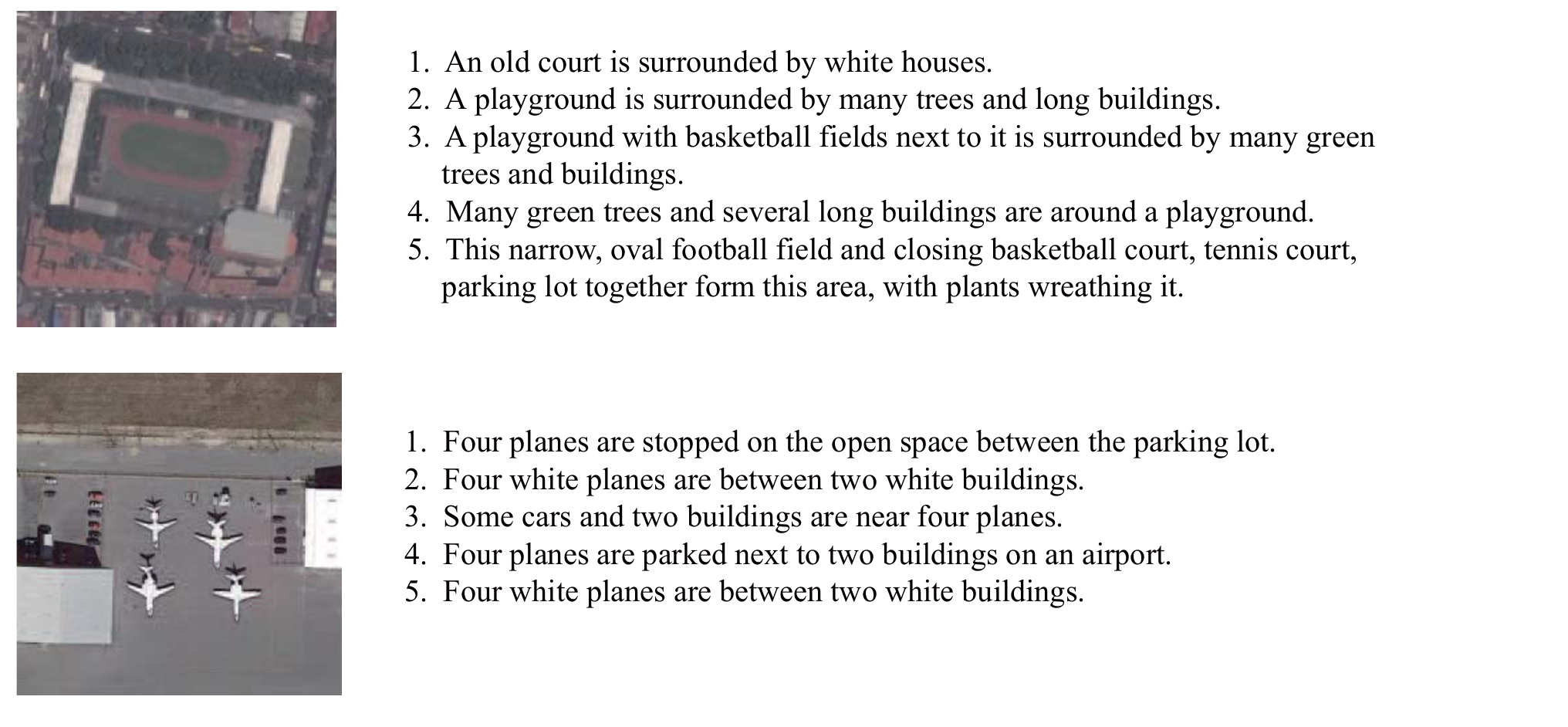}}
    \caption{The example of images and corresponding five sentences each image selected from our dataset.}
    \label{FigureExampleRISDC}
    \normalsize
\end{figure*}
As the images are captured from the airplane or satellite, there is no concept of direction, such as up, down, left, right, instead, we use `near' and `next to' to replace. In the course of formulating instructions, we take the work of natural image captioning \cite{Hodosh2013853,Young201467,Chen2015} as a reference. All the captions of RSICD are generated by volunteers with annotation experience and related knowledge of remote sensing field. We give five sentences to every remote sensing image. To prove the diversity, we let every volunteer to provide one or two sentences for a remote sensing image. And there are some annotated instructions as follows.
\begin{itemize}
\item Describe all the important parts of the remote sensing image.
\item Do not start the sentences with ``There is" when there are more than one object in an image.
\item Do not use the vague concept of words like large, tall, many, in the absence of contrast.
\item Do not use direction nouns, such as north, south, east and west.
\item The sentences should contain at least six words.
\end{itemize}

\begin{table*}[htb]
\centering
\caption{Number of images of each class in RSICD datasets (the total number of images is 10921)}
\label{NumberRSICD}
\scalebox{0.95}{
\begin{tabular}{|c|c|c|c|c|c|}
\hline
\hline
Class & Number&Class & Number&Class & Number\\
\hline
Airport&420 &Farmland&370&Playground&1031\\
\hline
Bare Land&310&Forest&250&Pond&420\\
\hline
Baseball Field&276&Industrial&390&Viaduct&420\\
\hline
Beach&400&Meadow&280&Port&389\\
\hline
Bridge&459&Medium Residential&290&Railway Station&260\\
\hline
Center&260&Mountain&340&Resort&290\\
\hline
Church&240&Park&350&River&410\\
\hline
Commercial&350&School&300&Sparse Residential&300\\
\hline
Dense Residential&410&Square&330&Storage Tanks&396\\
\hline
Desert&300&Parking&390&Stadium&290\\
\hline
\end{tabular}}
\end{table*}

The total number of sentences in RSICD is 24333, and the total words of these sentences are 3323. The detailed information on the annotations is as follows: 724 images are described by five different sentences, 1495 images are described by four different sentences, 2182 images are described by three different sentences, 1667 images are described by two different sentences and 4853 are described by one sentence. In order to make the sentences richer, we extended the sentences to 54605 sentences by duplicating randomly the existing sentences when there are not five different sentences to describe the same image.

\begin{figure*}[!t]
    \centering
    \scalebox{1.0}{\includegraphics{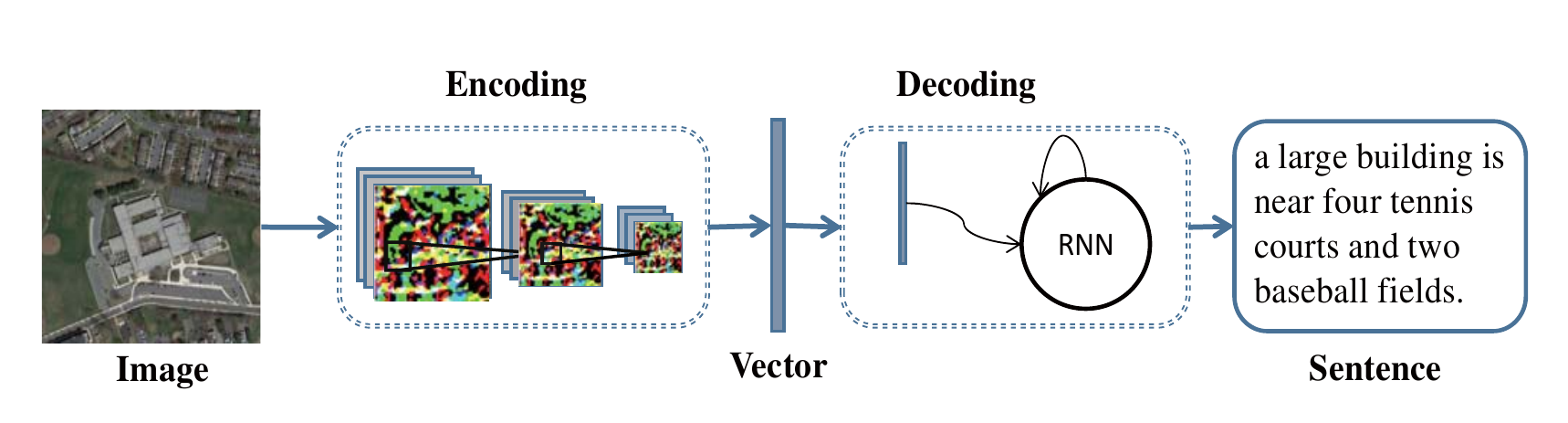}}
    \caption{The outline of encoder-decoder for remote sensing image caption: including training process and test process}
    \label{outline}
    \normalsize
\end{figure*}

Some example sentences of images are shown in Figure \ref{FigureExampleRISDC}. The detailed information of the dataset is shown in Table \ref{NumberRSICD}, and the object classes in the proposed dataset are determined by experts in remote sensing field. The relations of objects in showed images can be listed in ``near", ``in two rows", ``in", ``on", ``surrounded by", ``in two sides of".

\section{Remote sensing image captioning}
\label{sec:4}
\par
We present the outline of encoder-decoder (encoding an image to a vector, then decoding the vector to a sentence) for remote sensing image captioning task in Figure \ref{outline}.
First, we present remote sensing image representation methods (encoder). Then, the sentences generating models are introduced (decoder).

\subsection{Multimodal method}
The first method \cite{Qu2016124} utilizes a deep multimodal neural network to generate a coherent sentence for a remote sensing image. We introduce the method in three parts: representing remote sensing image, representing sentences and sentences generation.
\subsubsection{Representing remote sensing images}

To represent the remote sensing images, the existing image representation can be classified into two groups: handcrafted feature methods and learned feature methods. The handcrafted feature methods first extract the handcrafted features from each image and then obtain image representation by feature encoding techniques such as
\emph{Bag Of Words} (BOW) \cite{sivic2003video}, \emph{Fisher Vector} (FV) \cite{perronnin2010large} and \emph{Vector of Locally Aggregated Descriptors} (VLAD) \cite{Jegou20111704}. The learned deep feature methods can automatically learn the image representation from the training data via machine learning algorithms.

The performance of handcrafted feature methods strongly relies on the extraction of handcrafted local features and feature encoding methods. Lowe \emph{et al.} \cite{lowe2004distinctive} proposed a method to extract \emph{Scale-Invariant Feature Transform} (SIFT) features which are invariant to image scale change and rotation. BOW \cite{sivic2003video} represented a given image with the frequencies of local visual words, while FV \cite{perronnin2010large} uses gaussian mixture model  to encode the handcrafted local features. The VLAD \cite{Jegou20111704} concatenates the accumulation of the difference of the local feature vector and its cluster center vector.

Compared with the handcrafted features, deep features in computer vision have become more and more popular. The features extracted by CNNs have made great progress in many applications during the past several years \cite{Krizhevsky20121097,Jia2014675,Zeiler2014818,Simonyan2014,Szegedy20151}. We use fully connected layers of several CNNs, including AlexNet \cite{Krizhevsky20121097}, VGGNet \cite{Simonyan2014}, GoogLeNet \cite{Szegedy20151}, pre-trained on ImageNet dataset, to extract features of remote sensing images.
\begin{equation}
e_{0} = f_{FR}\left(I \right),
\end{equation}
where $I$ is a remote sensing image, $e_{0}$ is the feature of the remote sensing image whose dimension is notated $u$, and $f_{FR}$ is the feature representations process which the feature can be handcrafted feature or deep feature.

\subsubsection{Representing sentences}
In the first method, every word in a sentence is represented by a one-hot $K$ dimension word vector $w_{i}$, where $K$ is the size of the vocabulary. Then the word vector is projected to an embedding space by the embedding matrix $\textbf{E}_{s}$. $\textbf{E}_{s}$ is a $h \times K$ matrix, where $h$ is the dimension of embedding space. Sentence $y$ is encoded as a sequence of $h$ dimension projected word vectors \cite{rumelhart1988learning}:
\begin{equation}
w'_{i} = \textbf{E}_{s}\cdot w_{i} ,
\end{equation}
\begin{equation}
y = \{w'_{1},...,w'_{i},...,w'_{L}\},
\end{equation}

where $L$ is the length of the sentence.

\subsubsection{Sentences Generation}
In this subsection, a special \emph{Recurrent Neural Network} (RNN), called \emph{Long Short-Term Memory networks} (LSTM), is exploited to generate the sentences. This is because \emph{Long Short-Term Memory networks} is more complicated than original recurrent neural networks.

Sentence generating process is more like a ``human", which means that the human thoughts have persistence, and the next word can be predicted according to previous words to some extent. The \emph{Recurrent Neural Networks} (RNNs) \cite{williams1989learning} satisfy this property with loops allowing information to persist.

\begin{figure*}[!t]
    \centering
    \scalebox{0.6}{\includegraphics{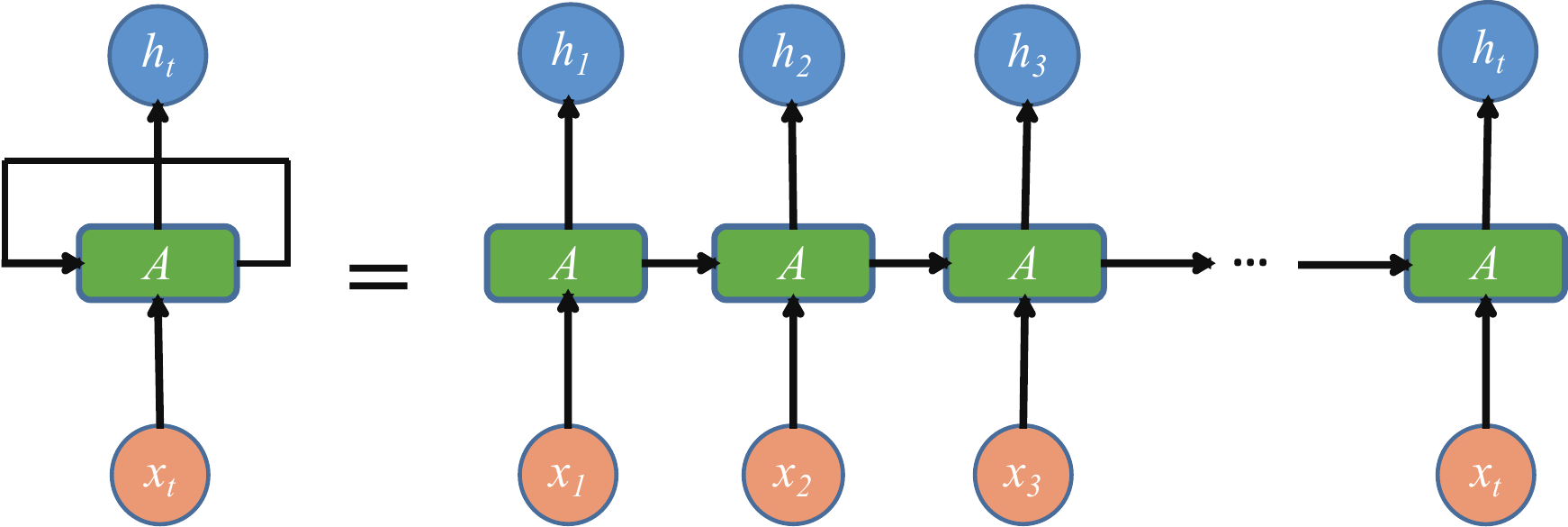}}
    \caption{ The structure of RNNs.}
    \label{rnnstructure}
    \normalsize
\end{figure*}

The structure of RNNs is shown in Figure \ref{rnnstructure}. As shown in Figure \ref{rnnstructure}, some input $x_{t}$, go through a neural network, $A$, and output a value $h_{t}$ (the range of $t$ is from 1 to $N$), where the $t$ is the state of RNNs. The information contained by the input is passed from one step to the next. The previous information contained by the state of RNNs can be passed to the present state is the main superiority of RNNs. But in some cases, the information that present task needs is far away from the current state. The problem of this long-term dependencies cannot be well solved with RNNs \cite{hochreiter1991untersuchungen}.

\begin{figure*}[!t]
    \centering
    \scalebox{0.6}{\includegraphics{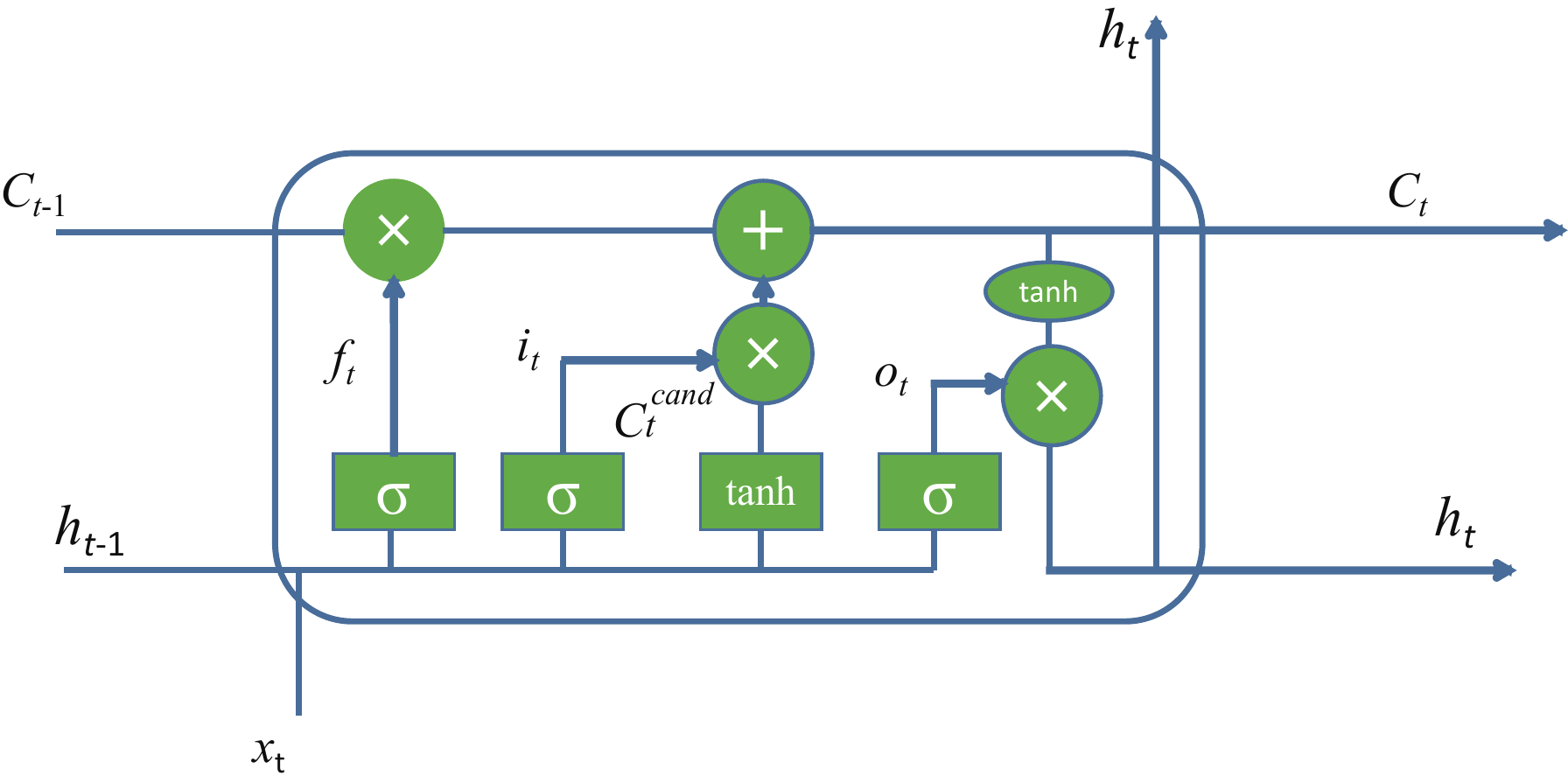}}
    \caption{ The structure of LSTM.}
    \label{lstm}
    \normalsize
\end{figure*}

To address the aforementioned problem, \emph{Long Short-Term Memory networks} (LSTM) is proposed in \cite{hochreiter1997long} to handle this long-term dependencies problem. LSTM is designed to address the long-dependency problem by using gates to control the information passed through the networks. The structure of LSTM is shown in Figure \ref{lstm}. The first step of LSTM is using a forget gate to decide what information to throw away. Then, an input gate decides the values we will update, and a tanh layer is used to generate candidate values that could be added to the state of networks. The output is the cell state filtered by an output gate.


The process of sentence generating is considered as a problem that maximizing the correct sentence generating probability conditioned on the given image information.

At training stage, the feature of the remote sensing image $I$ can be obtained by CNNs. Then the image feature and the corresponding sentences are fed into RNN or LSTM to train a model to predict word one by one given the image. The procedure of the training can be represented by the following formulations:

\begin{eqnarray}
h_{t} =
\begin{cases}
g\left(w'_{t} + \textbf{E}_{n}h_{t-1} + \textbf{E}_{m}e_{0}\right),& t = 1,\\
g\left(w'_{t} + \textbf{E}_{n}h_{t-1}\right),& t = 2,...,N,
\end{cases}
\end{eqnarray}
\begin{equation}
p{({w_{t + 1}})_i} = \frac{{exp ({{({\textbf{E}_o}{h_t})}_i})}}{{\sum\nolimits_{k=1}^{K + 1} {exp ({{({\textbf{E}_o}{h_t})}_k})} }}, i = 1,...,K+1,
\label{softmax}
\end{equation}
where $e_{0}$ is the feature of remote sensing image, $\textbf{E}_{n}\in\mathbb{R}^{{h} \times {h}}$, $\textbf{E}_{m}\in\mathbb{R}^{{h} \times {u}}$, $\textbf{E}_{o}\in\mathbb{R}^{\left({K+1}\right) \times {h}}$ are learnable parameters. The feature is imported only when $t = 1$ and the range of $t$ is from 1 to $N$. $g(\cdot)$ represents the process of the RNN or LSTM. $h_{t}$ is the output of state $t$ whose dimension is $h$. $w'_{t}$ means the corresponding words in the sentence $y$, and $w'_{1}$ and $w'_{N}$ are special token representing the START and END vector respectively. The final output $\textbf{E}_{o}h_{t}$ (including the END token, so the output dimension is $K+1$) goes through a softmax function \ref{softmax} to get the probability of the next word $p(w_{t+1})\in\mathbb{R}^{K+1}$.

And the best parameters of the the proposed model can be obtained at training stage by minimizing the following loss function:
\begin{equation}
loss(I,y) = - \sum_{t=1}^{N}{\rm log} p(w_{t}),
\end{equation}
At sentence generating stage, the feature of remote sensing image is fed into the model to generate word one by one forming a sentence.

\subsection{Attention based method}

The second method is an attention based model proposed in \cite{Vinyals20153156}. Two kinds of attention based manners are introduced including a deterministic manner and a stochastic manner. The deterministic manner uses standard backpropagation techniques to train the model, while the stochastic manner trains the model by maximizing a lower bound. In order to look different parts of an image, attention based method uses a different image representing method which is introduced as follows.

\subsubsection{Representing remote sensing images}

The features of lower convolution layers of CNNs represent the local feature compared with the fully connected layer \cite{zeiler2014visualizing}. In order to obtain the correspondence between the original remote sensing image and the feature vectors, the features of convolution layers are extracted in this method. For each remote sensing image, $M$ vectors are extracted and the dimension of each vector corresponding to a part of the remote sensing image is $D$.
\begin{equation}
a = \{\textbf{a}_{1},...,\textbf{a}_{M}\}, \textbf{a}_{i}\in\mathbb{R}^{D},
\end{equation}
where $a$ is called annotation vector contain a set of feature vectors.

\subsubsection{Representing sentences}
In the attention based method, a sentence $y$ is encoded as a sequence of 1-of-$K$ encoded words:
\begin{equation}
y = \{\textbf{y}_{1},...,\textbf{y}_{L}\},  \textbf{y}_{i}\in\mathbb{R}^{K},
\end{equation}
where $K$ is the size of the vocabulary and $L$ is the length of the sentence.
\subsubsection{Sentences Generation}
The key idea of attention based methods is to exploit the information of former state to decide ``where to look" in this state, and then the next word is predicted based on this information. The method of generating sentences in this subsection is LSTM.

The deterministic method, named as ``soft" attention, gives a weight to different parts of an image according to the information of former state to decide ``where to look". The stochastic method, named as ``hard" attention, uses a sampling strategy to look different parts of an image, then uses reinforcement learning to get an overall better result.

In order to generate sentences using LSTM, the inputs of LSTM in the second method are $y_{t-1}$ and $\hat{z_{t}}$ unlike in the first method which instead are $e_{0}$ and $w_{t}$. $\hat{z_{t}}$ is called context vector which is computed by different attention manners from the annotation vectors $a_{i}$.

The initial states of LSTM are predicted by an average of annotation vectors imported through two separate Multi-layer Perceptron (MLPs):
\begin{equation}
c_{0} = f_{init,c}\left(\frac{1}{M}\sum_{i}^{M}a_{i}\right),
\end{equation}
\begin{equation}
h_{0} = f_{init,h}\left(\frac{1}{M}\sum_{i}^{M}a_{i}\right),
\end{equation}
In the attention based method, a deep output layer is used to compute output word probability given the context vector, LSTM state and previous word:
\begin{equation}
p\left(\textbf{y}|\textbf{a},\textbf{y}_{1}^{t-1}\right)= {\rm exp}\left(\mathbf{L}_{o}y_{t-1}+\mathbf{L}_{h}h_{t}+\mathbf{L}_{z}\hat{z_{t}}\right),
\end{equation}
where $\mathbf{L}_{o}\in \mathbb{R}^{K \times m}$, $\mathbf{L}_{h}\in \mathbb{R}^{m \times n}$, $\mathbf{L}_{z}\in \mathbb{R}^{m\times D}$, and an embedding matrix $\mathbf{E}$ are parameters initialized randomly.

\section{Experiments}
\label{sec:5}
\par
We evaluate two methods mentioned before: multimodal method and attention based method. The feature used for multimodal method including deep CNNs representations (models are pre-trained on ImageNet dataset) and handcrafted representations such as SIFT, BOW, FV and VLAD. The datesets used in this section are introduced in Section \ref{sec:3}. Firstly, we introduce the metrics of remote sense image captioning. Secondly, we present the result of multimodal method. Then, the result of attention based method is given. Finally, experimental analysis is present to show that our dataset is more scientific than others.
\par
The experimental setup in this section: the word embedding dimension and hidden state dimension of RNN are respectively set to 256 and 256 for multimodal method, and the learning rate of multimodal method is 0.0001. The word embedding dimension and hidden state dimension of LSTM are respectively set to 512 and 512 for attention based method, and the learning rate of attention based method is 0.0001.

\subsection{Metrics of remote sensing image captioning }
The metrics used in this paper including BLEU \cite{Papineni2002311}, ROUGE\_L \cite{lin2004rouge}, METEOR \cite{lavie2014meteor}, CIDEr \cite{Vedantam20154566}. BLEU measures the co-occurrences of $n$-gram between the generated sentence and the reference sentence, where $n$-gram is a set of one or more ordered words. $n$ in this paper is 1, 2, 3 and 4. ROUGE\_L is a metric calculating F-measure given the length of the \emph{Longest Common Subsequence} (LCS). Unlike $n$-gram, the calculation of LCS includes the situation that there are other words between the $n$-gram. METEOR is computed by generating an alignment between the reference sentence and generated sentence. CIDEr measures the consensus by adding a \emph{Term Frequency Inverse Document Frequency} (TF-IDF) weighting for every $n$-gram. The CIDEr metric has more reference value compared with BLEU, ROUGE\_L, METEOR \cite{Vedantam20154566}.

The range of the bleu\-1, bleu\-2, bleu\-3, bleu\-4, METEOR and ROUGE\_L is between 0 and 1. And the closer to 1 the metrics are, the more likely the generated sentence is to the reference sentences in dataset. The range of CIDEr in tables \ref{resUCMmulti}, \ref{resSydneymulti}, \ref{resRSICDmulti}, \ref{diffCNNsLSTM}, \ref{differCNNsUCMattention}, \ref{differCNNsSydneyhattention}, \ref{differCNNsRSICDattention}, \ref{generalization} is between 0 and 5. The bigger the CIDEr is, the better the generated sentences are.

Except for objective metrics mentioned above, subjective metrics are also given to better understand the quality of generated sentences and the generalization capabilities of models trained on different datasets in Section \ref{analysis:2}.
\subsection{The results of multimodal method}
\begin{table*}[htb]
\centering
\caption{Result of multimodal method on UCM-captions dataset}
\label{resUCMmulti}
\scalebox{0.95}{
\begin{tabular}{|c|c|c|c|c|c|c|c|c|c|}
\hline
\hline
& & bleu\-1 & bleu\-2 & bleu\-3 & bleu\-4 & METEOR & ROUGE\_L & CIDEr\\
\hline
\multirow{4}{*}{RNN}& SIFT &0.57303&	0.44354	&0.37963&	 0.33883&0.24611	 &0.53344&1.70329\\
\cline{2-9}
& BOW& 0.41067	&0.22494&	0.14519	 &0.10951	 &0.10976	 &0.34394	 &0.30717\\
\cline{2-9}
 &FV &0.5908&0.46026&0.39681&0.35446&0.25975&0.54575&1.68732\\
\cline{2-9}
 &VLAD &\textbf{0.63108}&\textbf{0.51928}&\textbf{0.46057} &\textbf{0.42087} &\textbf{0.2971}&\textbf{0.5878}&\textbf{2.00655}\\
\hline
\multirow{4}{*}{LSTM}& SIFT & 0.55168&0.41656	&0.34891&	 0.30403&	 0.24324&	 0.52354&	 1.36033\\
\cline{2-9}
& BOW&0.39109	&0.18767	&0.10892	 &0.07058&	 0.1315	 &0.33136	 &0.17812\\
\cline{2-9}
& FV &0.58972	&0.46678	&0.40799	 &0.36832	 &0.26975	 &0.5595	 &1.84382\\
\cline{2-9}
& VLAD &\textbf{0.70159}&\textbf{0.60853}&\textbf{0.54961}	 &\textbf{0.50302}&\textbf{0.34635}&\textbf{0.65197}&\textbf{2.31314}\\
\hline
\end{tabular}}
\end{table*}

\begin{table*}[htb]
\centering
\caption{Result of multimodal method on Sydney-captions dataset}
\label{resSydneymulti}
\scalebox{0.95}{
\begin{tabular}{|c|c|c|c|c|c|c|c|c|c|}
\hline
\hline
& & bleu\-1 & bleu\-2 & bleu\-3 & bleu\-4 & METEOR & ROUGE\_L & CIDEr\\
\hline
\multirow{4}{*}{RNN}& SIFT &0.58891 &0.48179 &0.42676	 &0.38935 &0.26162 &0.53923 &\textbf{1.21719}\\
\cline{2-9}
& BOW&0.53103&	0.40756	&0.33192	 &0.2788&	 0.24905	 &0.49218	 &0.70189\\
\cline{2-9}
& FV &\textbf{0.60545}& \textbf{0.49113}& \textbf{0.4259}&	 \textbf{0.37855}& \textbf{0.27578}& \textbf{0.55402}& 1.1777\\
\cline{2-9}
& VLAD &0.56581&0.45141	&0.38065	 &0.32787	 &0.26718	 &0.52706	 &0.93724\\
\hline
\multirow{4}{*}{LSTM}& SIFT &0.57931	&0.47741&	 0.41828	 &0.37400&	 0.27072&	 0.53660&	 0.98730\\
\cline{2-9}
& BOW&0.53103	&0.40756	 &0.33192&	0.2788&	 0.24905	 &0.49218	 &0.70189\\
\cline{2-9}
& FV &\textbf{0.63312} &\textbf{0.53325}&\textbf{0.47352}	 &\textbf{0.43031}&	 \textbf{0.29672}& \textbf{0.57937}	 &\textbf{1.47605}\\
\cline{2-9}
& VLAD  &0.49129&0.34715&0.27598&	 0.23144&	 0.19295&	 0.42009&	 0.91644\\
\hline
\end{tabular}}
\end{table*}

\begin{table*}[htb]
\centering
\caption{Result of multimodal method on RSICD dataset}
\label{resRSICDmulti}
\scalebox{0.95}{
\begin{tabular}{|c|c|c|c|c|c|c|c|c|c|}
\hline
\hline
& & bleu\-1 & bleu\-2 & bleu\-3 & bleu\-4 & METEOR & ROUGE\_L & CIDEr\\
\hline
\multirow{4}{*}{RNN}& SIFT & 0.47652	&0.28275	 &0.19611&	 0.14558	 &0.18814&	 0.39913	 &0.78827  \\
\cline{2-9}
& BOW&0.44007	&0.23826	&0.15138	&0.1041	&0.16844	 &0.3605	 &0.46671\\
\cline{2-9}
& FV &0.48591	&0.30325&	0.21861&	 \textbf{0.16779}	 &0.19662	 &0.41742	 &\textbf{1.05284}\\
\cline{2-9}
& VLAD  &  \textbf{0.49377}&\textbf{0.30914}	 &\textbf{0.22091}	 &0.16767	 &\textbf{0.19955}	 &\textbf{0.42417}&	 1.03918\\
\hline
\multirow{4}{*}{LSTM}& SIFT &0.48591	&0.30325&	 0.21861&	 0.16779	 &0.19662	 &0.41742	 &1.05284\\
\cline{2-9}
& BOW&0.48176&	0.29082&	0.20423&	0.15334&	 0.19436&	 0.39948&	 0.91521\\
\cline{2-9}
& FV& 0.43418&0.24527&0.16345&0.11746&0.17115&0.38181&0.65306\\
\cline{2-9}
& VLAD &\textbf{0.50037}	&\textbf{0.3195}	 &\textbf{0.23193}	 &\textbf{0.17777}	&\textbf{0.20459}	 &\textbf{0.43338}	 &\textbf{1.18011}\\
\hline
\end{tabular}}
\end{table*}
In this subsection, we evaluate multimodal method based on different kinds of features with randomly 80\% for training, 10\% for validation and 10\% for test on UCM-captions dataset \cite{Qu2016124}, Sydney-captions \cite{Qu2016124} datasets and our dataset RSICD. We firstly test four handcrafted representations for captioning, and then use the different CNNs.

\subsubsection{Results based on handcrafted representations}
To evaluate the generated sentences based on handcrafted representations, four handcrafted representations are conducted including SIFT, BOW, FV and VLAD. Each remote sensing image sized 224$\times$224 is segmented evenly to sixteen patches sized 56$\times$56. For each patch, a SIFT feature is obtained by \emph{Principal Component Analysis} (PCA) of the origin SIFT features. Finally, sixteen SIFT features are concatenated into a vector to represent the image. Other handcrafted representations are based on SIFT. For BOW representation, the dictionary size is 1000.  Tables \ref{resUCMmulti}-\ref{resRSICDmulti} illustrate that the result of LSTM is better than that of RNN on UCM-captions dataset and RSICD dataset. Since the LSTM solves the long-term dependencies problem of RNNs, the sentence generated by LSTM can better depict a remote sensing image than the one generated by RNNs \cite{hochreiter1997long}. For all four handcrafted representations, VLAD performs the best on UCM-captions dataset and RSICD dataset. The result of LSTM on Sydney-captions is not as good as RNN. But according to \cite{schmidhuber2002learning}, LSTM should outperform RNN. This is perhaps caused by the imbalance of Sydney-captions and the further analysis is presented in Section \ref{analysis:1}.

\begin{table}[htb]
\centering
\caption{Number of images of each class in Sydney-captions dataset}
\label{numberclasssydney}
\scalebox{0.95}{
\begin{tabular}{|c|c|}
\hline
\hline
Class & Number\\
\hline
Residential & 242\\
\hline
Airport & 22\\
\hline
Meadow & 50 \\
\hline
River & 45 \\
\hline
Ocean & 92 \\
\hline
Industrial & 96 \\
\hline
Runway & 66\\
\hline
Total & 613 \\
\hline
\end{tabular}}
\end{table}

\begin{table*}[htb]
\centering
\caption{Results of multimodal method using different CNNs on RSICD using LSTM}
\label{diffCNNsLSTM}
\scalebox{0.95}{
\begin{tabular}{|c|c|c|c|c|c|c|c|c|}
\hline
\hline
CNN& bleu\-1 & bleu\-2 & bleu\-3 & bleu\-4 & METEOR & ROUGE\_L & CIDEr\\
\hline
VGG19 &\textbf{0.58330}&	\textbf{0.42259}&	 \textbf{0.33098}&	 \textbf{0.27022}&\textbf{0.26133}&	 0.51891&	 2.03324\\
\hline
VGG16 &0.57951	&0.41729	&0.32462	 &0.26358	 &0.26116	 &0.51811	 &2.01663\\
\hline
AlexNet &   0.57905&	0.41871	&0.32628	 &0.26552&	 0.26103	 &\textbf{0.51913}	 &\textbf{2.05261}\\
\hline
GoogLeNet &0.57847&0.41363	&0.32042	 &0.2595&	0.25856	 &0.51318	 &2.00075 \\
\hline
\end{tabular}}
\end{table*}

\subsubsection{Results based on different CNNs}
In order to evaluate the generated sentences based on CNNs features, the experiments based on CNNs features are conducted in this subsection. The experiments of different CNNs features on dataset UCM-captions and Sydney-captions have been done in \cite{Qu2016124}, and the result on our dataset RSICD is shown in Table \ref{diffCNNsLSTM}. The result shows that all the CNNs features are better than handcrafted features. All the CNNs features get almost the same result. In CNNs features, AlexNet gets the best result on ROUGE\_L and CIDEr and VGG19 gets the best result on other objective metrics with a little superiority than others. This means that representation of CNNs features are powerful for remote sensing image captioning task.

\subsubsection{Results of different training ratios}
\begin{figure*}[!t]
    \centering
    \scalebox{0.445}{\includegraphics{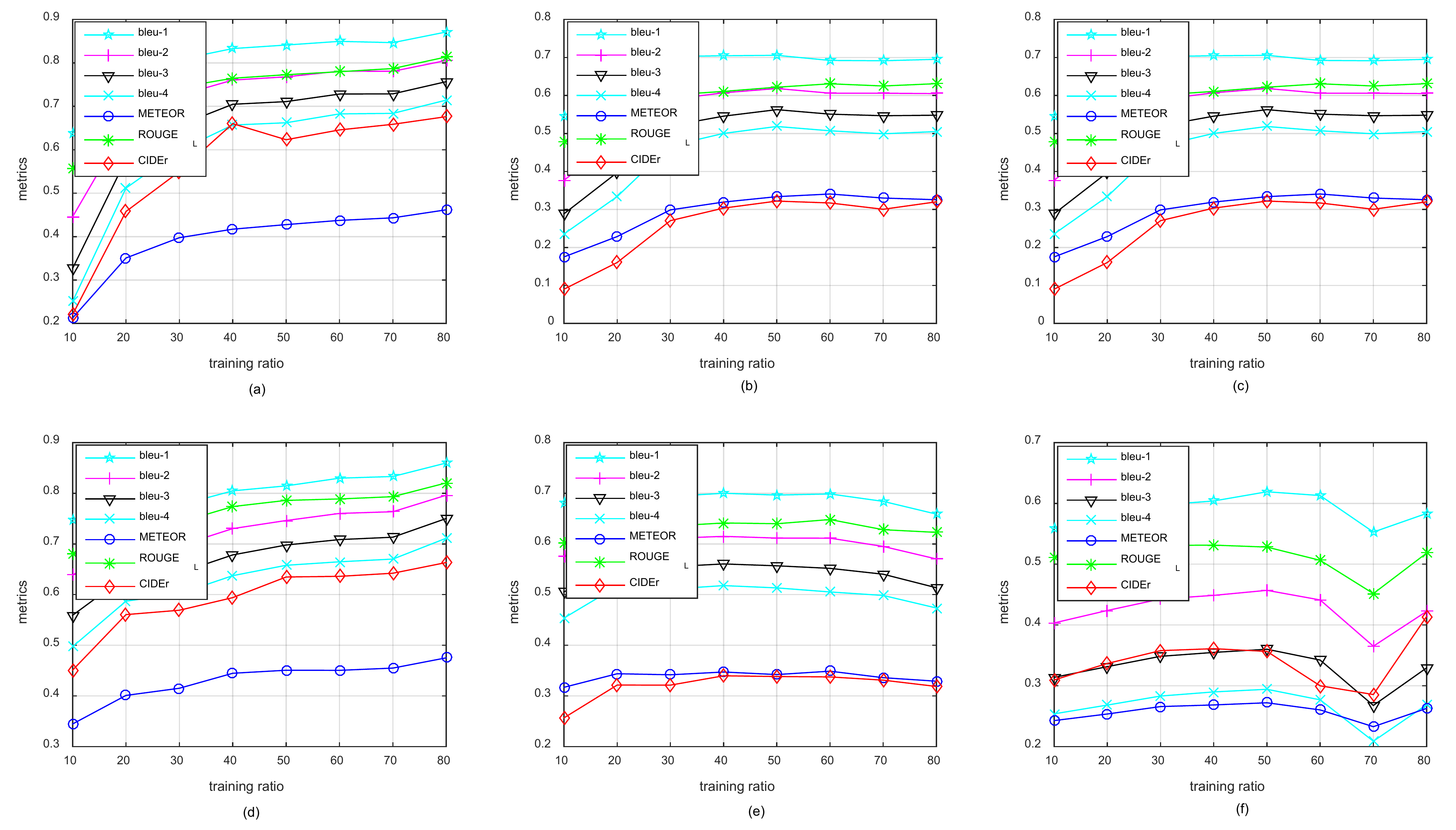}}
    \caption{(a) Metrics of multimodal method using RNN on UCM-captions dataset. (b) Metrics of multimodal method using RNN on Sydney-captions dataset. (c) Metrics of multimodal method using RNN on RSICD dataset. (d) Metrics of multimodal method using LSTM on UCM-captions dataset. (d) Metrics of multimodal method using LSTM on Sydney-captions dataset. (f) Metrics of multimodal method using LSTM on RSICD dataset.}
    \label{ratioall}
    \normalsize
\end{figure*}

In order to study the influence of the ratio of training images to the caption result, We use 10\% images of dataset for validation, and change the ratio of training and testing images to observe the result based on VGG16 CNNs features. The results of different sentence generating methods based on different datasets are shown in Figure \ref{ratioall}. The abscissa represents the proportion of training samples, and the ordinate is the corresponding result. The different metrics are shown in different colors and shapes.

\begin{table*}[!htb]
\centering
\caption{ Result of attention based method using different CNNs on UCM-captions dataset}
\label{differCNNsUCMattention}
\scalebox{0.95}{
\begin{tabular}{|c|c|c|c|c|c|c|c|c|}
\hline
\hline
CNN &  model & bleu\-1 & bleu\-2 & bleu\-3 & bleu\-4 & METEOR & ROUGE\_L & CIDEr\\
\hline
\multirow{2}{*}{VGG19} & soft & 0.74569 & 0.65976 &	0.59485 & 0.53932	 & 0.39629&	 0.72421	 &2.62906\\
\cline{2-9}
 & hard & 0.78493&	0.70892&	0.65387&	 0.60604&	 0.42760&	 0.76564&	 2.84756\\
\hline
\multirow{2}{*}{VGG16} & soft & 0.7454&	 0.6545&	 0.58553&	 0.52502&	 0.38857&	 0.72368&	 2.61236\\
\cline{2-9}
 & hard &0.81571&	0.73124&	0.67024&	 0.61816&	 0.42630&	 0.76975&	 2.99472\\
\hline
\multirow{2}{*}{AlexNet} & soft &0.79693&	 0.71345&	 0.6514&	 0.59895&	 0.41676&	 0.74952&	 2.82368\\
\cline{2-9}
 & hard &0.78498&	0.70929&	0.65182&	 0.60167&	 0.43058&	 0.77357&	 3.01696\\
\hline
\multirow{2}{*}{GoogLeNet}&soft &0.76356&	 0.67662&	 0.61034&	 0.55371&	 0.40103&	 0.73996&	 2.85671\\
\cline{2-9}
 & hard &\textbf{0.83751}&	\textbf{0.76217}&	 \textbf{0.70420}&	 \textbf{0.65624}&	 \textbf{0.44887}&	 \textbf{0.79621}&	 \textbf{3.2001}\\
\hline
\end{tabular}}
\end{table*}

\begin{figure}[!t]
    \centering
    \scalebox{0.545}{\includegraphics{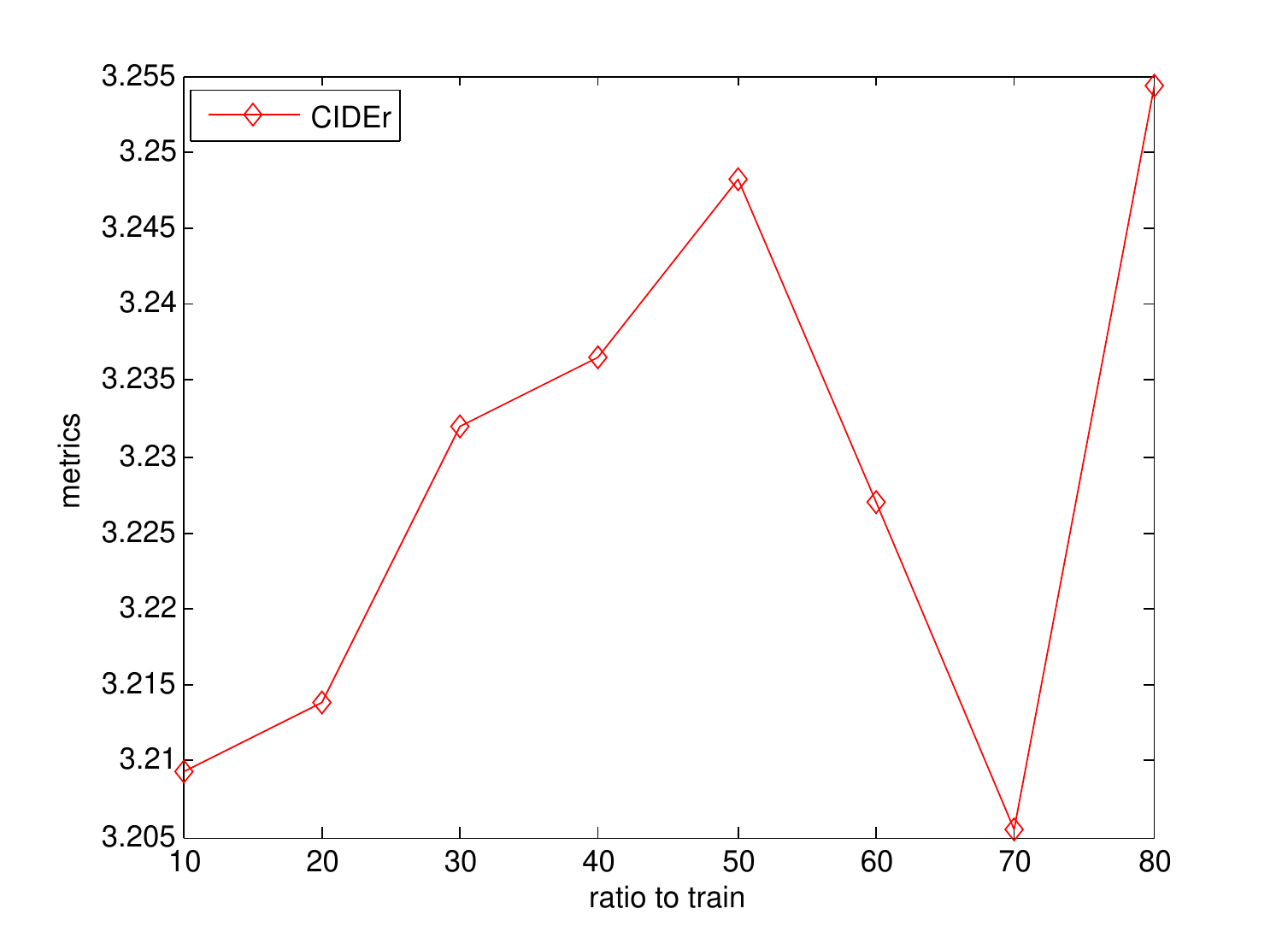}}
    \caption{Metrics of reference sentences on RSICD dataset.}
    \label{FigureRSICDmetricrefer}
    \normalsize
\end{figure}

As shown in Figure \ref{ratioall}, the results of all metrics are getting higher following the increasement of the training ratio on UCM-captions dataset \cite{Xia2016}. But as for Sydney-captions dataset in Figures \ref{ratioall} (b) and \ref{ratioall} (e), the metrics increase at first and then keep an almost stable result. This is caused by the imbalanced of the Sydney-captions dataset. As aforementioned, the most generated sentences on Sydney-captions dataset are related to ``residential area". From Figures \ref{ratioall} (c) and \ref{ratioall} (f), we can find that the performance is getting better following the increase of the training ratio on RSICD firstly. Then a fluctuation occurs when the training ratio is between 60\% to 80\%. The occurrence of this fluctuation is because the some sentences in RSICD are obtained by duplicating the existing sentences. In order to evaluate the reference sentences of RSICD, the most reliable metric CIDEr is used. As shown in Figure \ref{FigureRSICDmetricrefer}, the trend of the CIDEr of reference sentences is the same as the trend in Figures \ref{ratioall} (c) and \ref{ratioall} (f).

\subsection{The results of attention based method}
\begin{table*}[!htb]
\centering
\caption{ Result of attention based method using different CNNs on Sydney-captions dataset}
\label{differCNNsSydneyhattention}
\scalebox{0.95}{
\begin{tabular}{|c|c|c|c|c|c|c|c|c|}
\hline
\hline
CNN &  model & bleu\-1 & bleu\-2 & bleu\-3 & bleu\-4 & METEOR & ROUGE\_L & CIDEr\\
\hline
\multirow{2}{*}{VGG19} & soft & 0.72818& 0.63847& 0.56323	 &0.50056 &0.38699	 &0.71483	&2.11849\\
\cline{2-9}
 & hard & 0.73876&	0.63988&	0.56413	 &0.50223	 &0.37551&	 0.69795	 &2.04721\\
\hline
\multirow{2}{*}{VGG16} & soft &0.73216&	 \textbf{0.66744}&	 \textbf{0.62226}&\textbf{0.58202}&0.3942&	 0.71271&	 \textbf{2.4993}\\
\cline{2-9}
 & hard &0.75907&	0.66095&	0.58894&	 0.52582&	 0.38977&	 0.71885&	 2.18186\\
\hline
\multirow{2}{*}{AlexNet} & soft &0.74128 &0.65842	 &0.59043	 &0.53139	 &\textbf{0.39634}	 &\textbf{0.7214}	&2.12846\\
\cline{2-9}
 & hard &0.74082 &0.65373	&0.58528	 &0.52538	 &0.37027	 &0.69774	 &2.19594\\
\hline
\multirow{2}{*}{GoogLeNet}&soft &0.71284&	 0.62387&	 0.55274&	 0.4924	 &0.36746	 &0.6913	 &2.03435\\
\cline{2-9}
 & hard &\textbf{0.76893}& 0.66125&	0.58399	&0.51699	 &0.37193	 &0.68421	 &1.9863\\
\hline
\end{tabular}}
\end{table*}
\begin{figure}[!t]
    \centering
    \scalebox{0.545}{\includegraphics{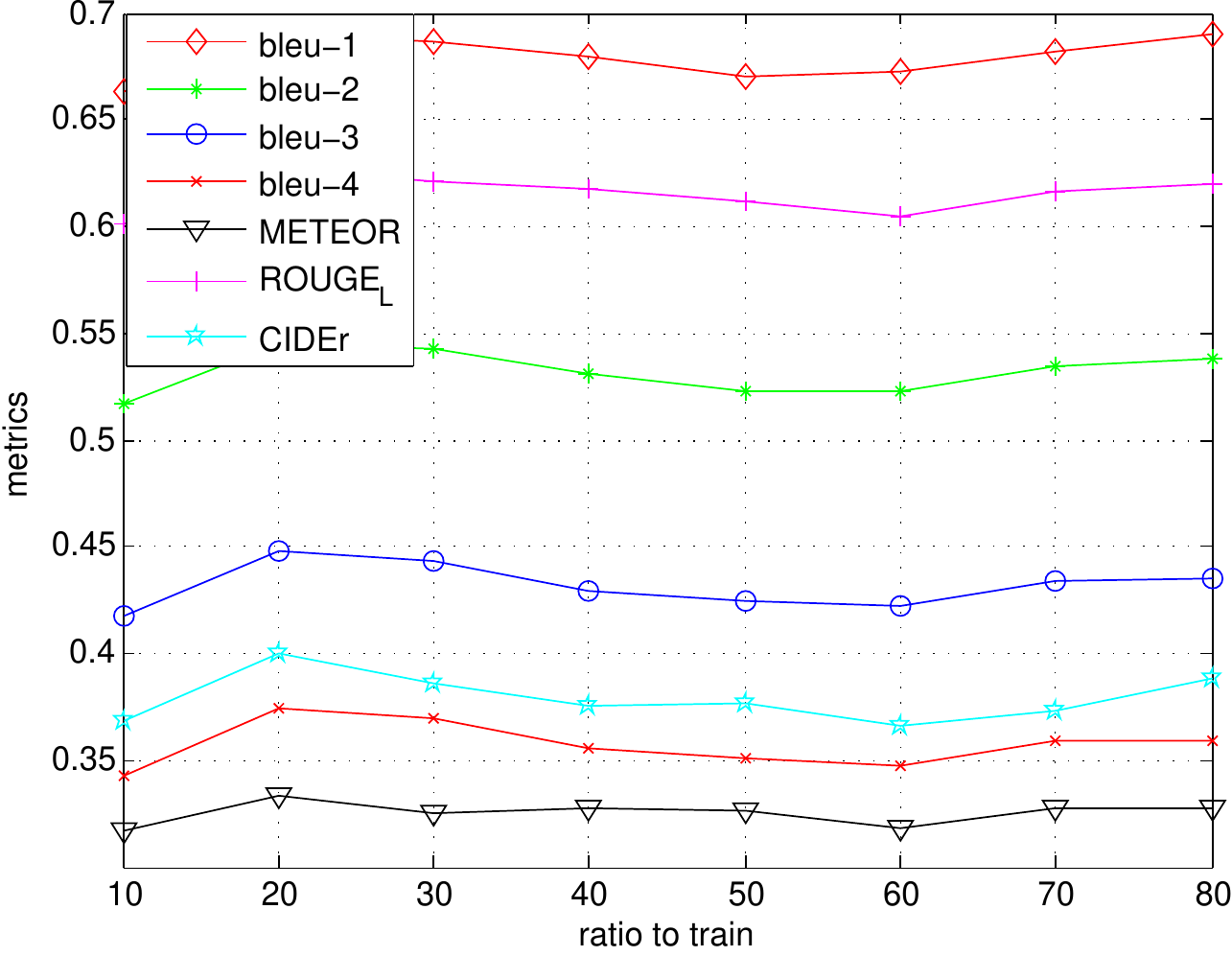}}
    \caption{Metrics of attention based method using LSTM on RSICD dataset.}
    \label{FigureRSICDattentionratio}
    \normalsize
\end{figure}

\begin{table*}[!htb]
\centering
\caption{ Result of attention based method using different CNNs on RSICD dataset}
\label{differCNNsRSICDattention}
\scalebox{0.95}{
\begin{tabular}{|c|c|c|c|c|c|c|c|c|}
\hline
\hline
CNN &  model & bleu\-1 & bleu\-2 & bleu\-3 & bleu\-4 & METEOR & ROUGE\_L & CIDEr\\
\hline
\multirow{2}{*}{VGG19} & soft &0.64633	 &0.50718	 &0.41137&	 0.33959	 &0.33332	 &0.61632&	 1.81861\\
\cline{2-9}
 & hard & 0.67925	&0.53052	&0.4296	 &0.35594	&0.3286	 &0.61769	 &1.89538\\
\hline
\multirow{2}{*}{VGG16} & soft &0.67534	 &0.53084&	0.43326	 &0.36168	 &0.32554	 &0.61089&	1.96432\\
\cline{2-9}
 & hard &0.66685	&0.51815	&0.4164&	 0.34066	 &0.32007	 &0.6084	 &1.79251\\
\hline
\multirow{2}{*}{AlexNet} & soft &0.65638&	 0.51489&	 0.41764&	 0.34464	 &0.32924	 &0.61039&	 1.87415\\
\cline{2-9}
 & hard &\textbf{0.68968}&	0.5446	&0.44396	 &0.36895	 &\textbf{0.33521}	 &0.62673	 &1.98312\\
\hline
\multirow{2}{*}{GoogLeNet}&soft &0.67371&	 0.5303	 &0.43237	 &0.35982	 &0.33389	 &0.62119	 &1.96519\\
\cline{2-9}
 & hard &0.68813&	\textbf{0.54523}	&\textbf{0.44701}	 &\textbf{0.3725}	 &0.33224	 &\textbf{0.62837}	 &\textbf{2.02145}\\
\hline
\end{tabular}}
\end{table*}
\begin{figure*}[!t]
    \centering
    \scalebox{0.845}{\includegraphics{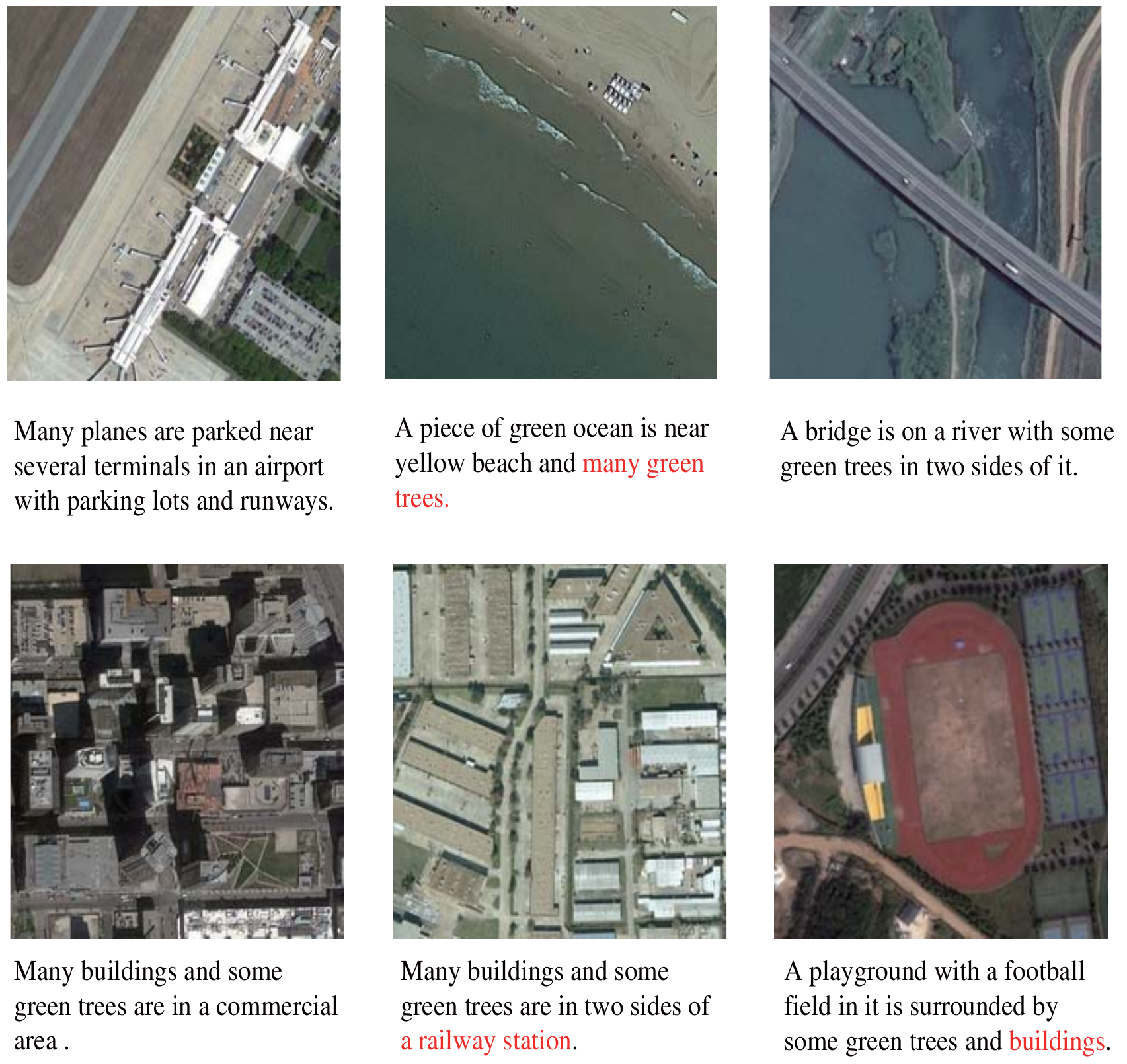}}
    \caption{The result of image captioning on RSICD dataset.}
    \label{exampleresult}
    \normalsize
\end{figure*}

In this subsection, the results of attention based method are discussed.
\subsubsection{Results based on different CNNs}
In order to evaluate the attention based method based on different CNNs, experiments are conducted below. Since the attention based method is based on the convolutional feature of CNNs, the features used in attention based method are all convolutional features extracted by different CNN models. Specifically, for VGG16, the feature maps of \emph{conv5}\_\emph{3} sized 14$\times$14$\times$512 are used; for VGG19, the feature maps of \emph{conv5}\_\emph{4} sized 14$\times$14$\times$512 are used; for AlexNet, the feature maps of \emph{conv5} sized 13$\times$13$\times$256 are used; for GoogLeNet, the feature maps of \emph{inception}\_\emph{4c/3$\times$3} sized 14$\times$14$\times$512 are used. The results of convolutional features extracted by different models are shown in Tables \ref{differCNNsUCMattention}-\ref{differCNNsRSICDattention}. We can see that the results of ``hard" attention mechanism are better than results of ``soft" attention mechanism in most conditions. The ``hard" attention mechanism based on the convolutional features extracted by GoogLeNet gets the best result on UCM-captions dataset and RSICD dataset. But for Sydney-captions dataset, the result of ``soft" attention mechanism based on the convolutional feature extracted by VGG16 gets the best result.

\subsubsection{Results of different training ratios}
To evaluate the influence of different training ratios, the training ratio is changed to get different results. The results of attention based ``hard" method of different training ratios are shown in Figure \ref{FigureRSICDattentionratio}. It can be found that the metrics firstly increase and then keep an almost stable result. Features extracted from convolutional layer of AlexNet can represent exhaustive content of remote sensing images. As shown in Figure \ref{FigureRSICDattentionratio}, the metrics almost keep stable when the training ratio is more than 20\%. This is because the representation capability of AlexNet for remote sensing image captioning task is powerful.

Figure \ref{exampleresult} shows some remote sensing image captioning results. Most of the sentences generated describe the image accurately. Some wrong places are shown in red color. In fact, the wrong places of generated sentences are related to the images. For example, the shape of many buildings in the second image of the second row in Figure \ref{exampleresult} is like a railway station. The other reason of wrong places of generated sentences is the high co-occurrences of two words in dataset. For example, the ``trees" and ``buildings" are high frequency co-occurrences words in dataset, so the generated sentence of the third image of the second row in Figure \ref{exampleresult} includes ``buildings" even though there is no building in the image.
\subsection{Experimental analysis}
To further validate the proposed dataset, two sets of experiments are conducted in this subsection. The imbalance of Sydney-captions datasets is analyzed firstly. Then, human subjective evaluation is performed to compare the generalization capabilities of models trained on different datasets.

\subsubsection{The imbalance of Sydney-captions}
\label{analysis:1}
\begin{figure}[!t]
    \centering
    \scalebox{0.545}{\includegraphics{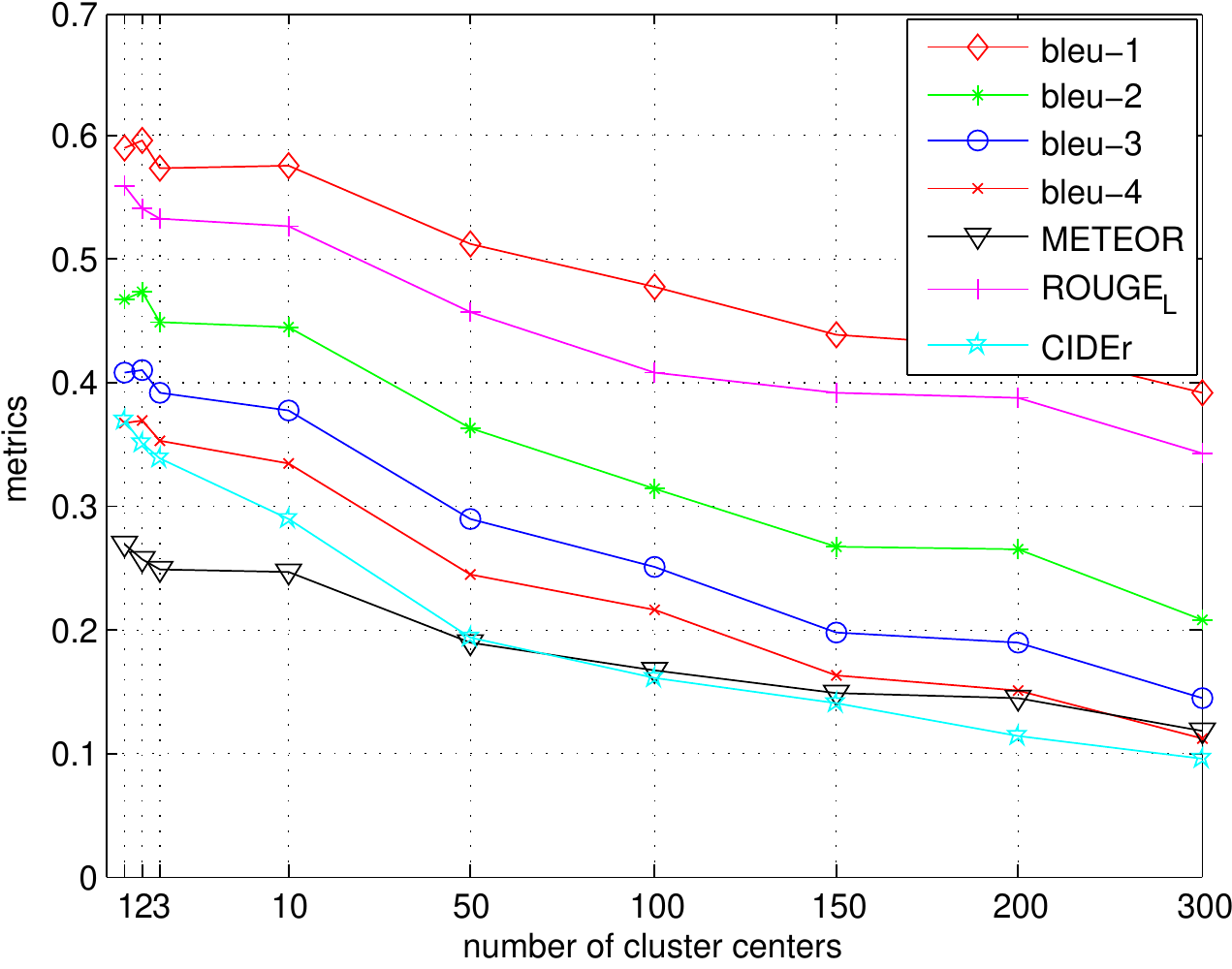}}
    \caption{The relations between metrics of multimodal method using LSTM with number of cluster centers of FV feature on UCM-captions dataset.}
    \label{Figureucmfv}
    \normalsize
\end{figure}

\begin{figure}[!t]
    \centering
    \scalebox{0.545}{\includegraphics{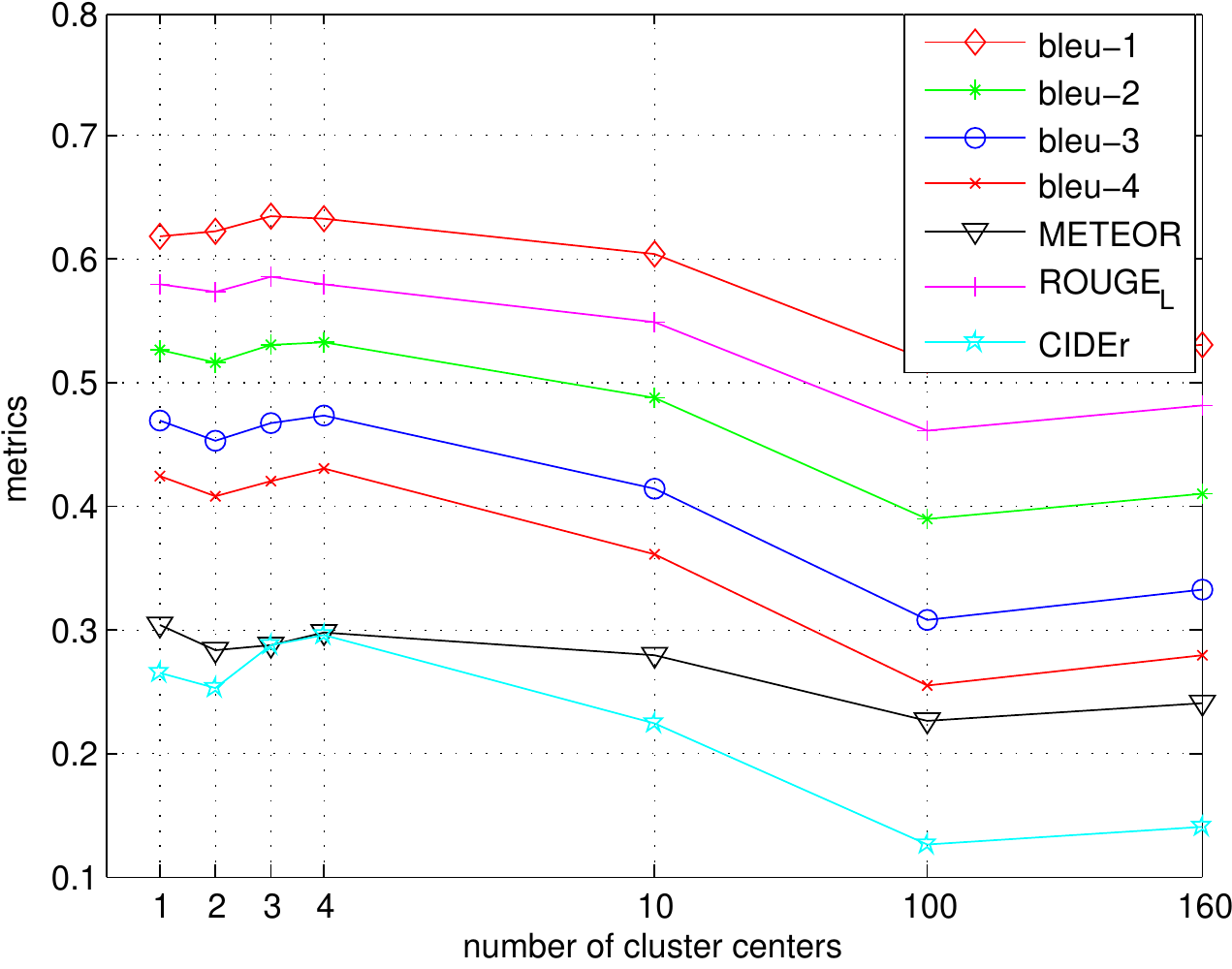}}
    \caption{The relations between metrics of multimodal method using LSTM with number of cluster centers of FV feature on Sydney-captions dataset.}
    \label{FigureSydneyfv}
    \normalsize
\end{figure}

\begin{figure}[!t]
    \centering
    \scalebox{0.545}{\includegraphics{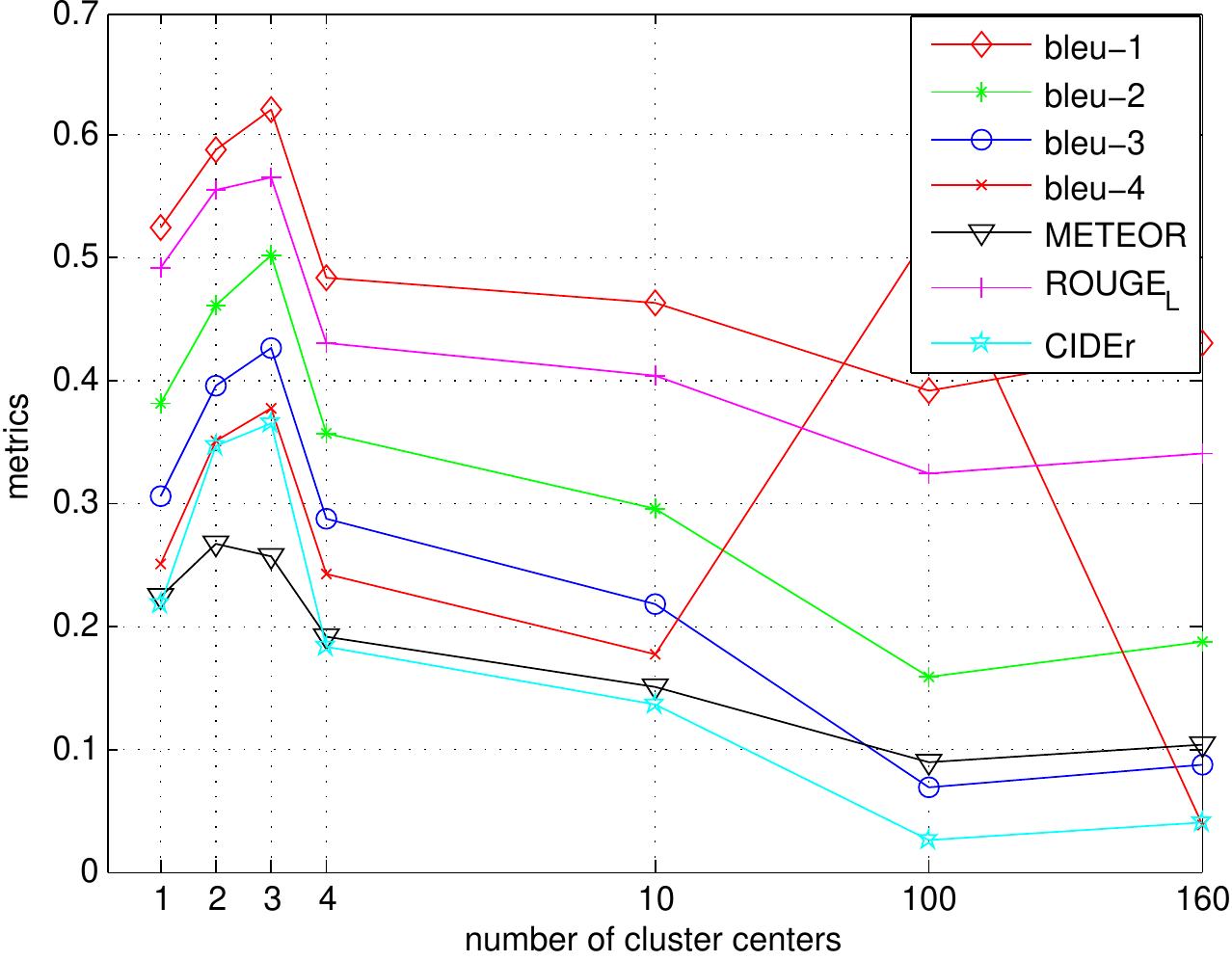}}
    \caption{The relations between metrics of multimodal method using LSTM with number of cluster centers of FV feature on balanced Sydney-captions dataset.}
    \label{FigureSydneyfvbalanced}
    \normalsize
\end{figure}

\begin{figure}[!t]
    \centering
    \scalebox{0.545}{\includegraphics{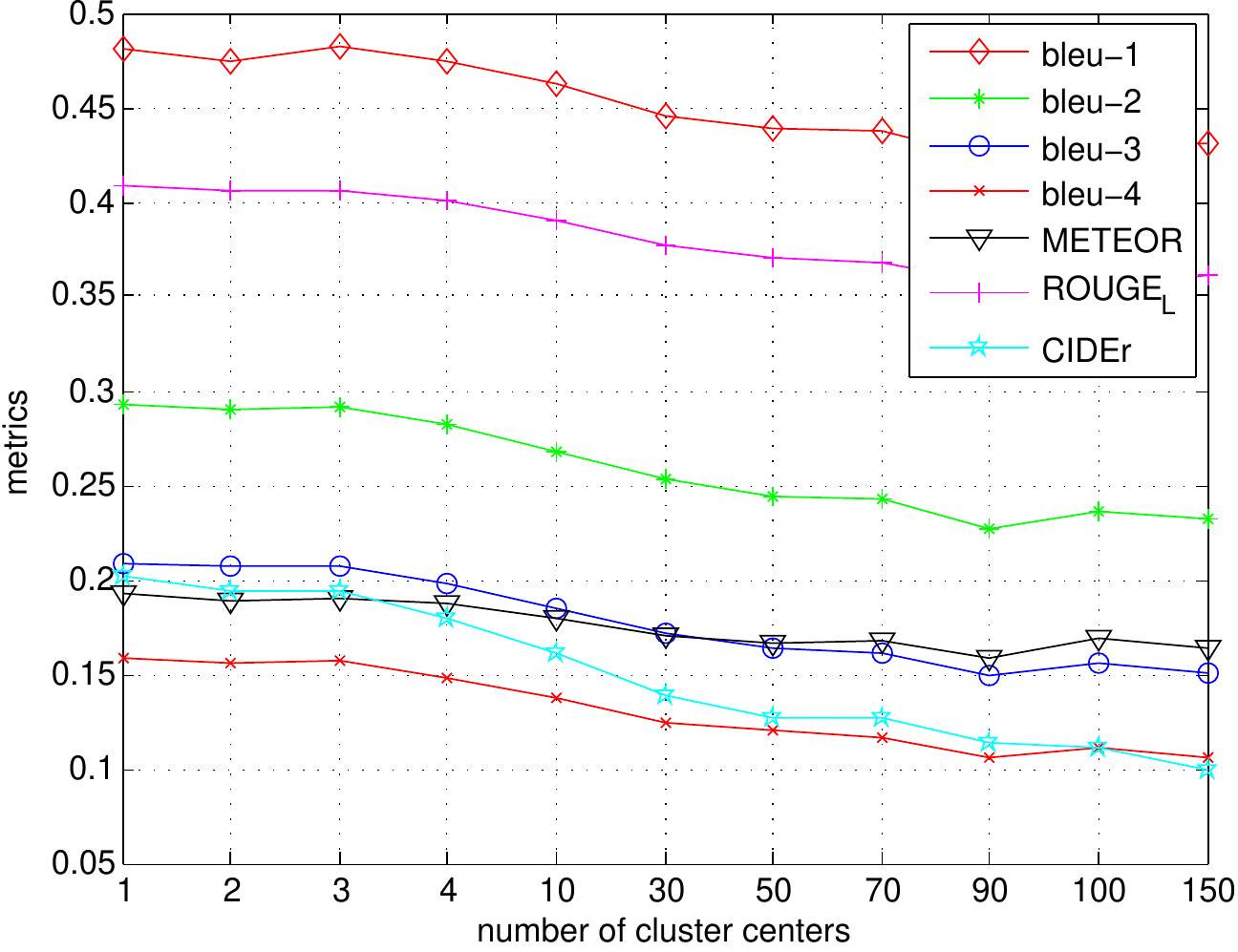}}
    \caption{The relations between metrics of multimodal method using LSTM with number of cluster centers of FV feature on RSICD dataset.}
    \label{FigureRSICDfv}
    \normalsize
\end{figure}

To verify the influence of unbalance of different kinds image numbers, we present the result of FV using different numbers of cluster center as the results of the FV is related to the number of cluster centers to construct a \emph{Gaussian Mixture Model} (GMM). The relations between the metrics with number of cluster centers of FV are shown in Figs. \ref{Figureucmfv}, \ref{FigureSydneyfv}, \ref{FigureSydneyfvbalanced} and \ref{FigureRSICDfv}. As shown in Figs. \ref{Figureucmfv} and \ref{FigureRSICDfv}, the metrics almost decrease with the increase of the number of center centers of FV. From the result of multimodal method based on FV of Sydney-captions dataset in Figure \ref{FigureSydneyfv}, it can be seen that the change of the metrics scores is not significantly related to cluster center numbers when the number of cluster centers are in the range of 1 to 10. We look through the generated sentences and find that almost all the generated sentences are related to ``residential area". After considering the imbalanced of the dataset, as shown in Table \ref{numberclasssydney}, the reason is the imbalance of data between different classes of images. The number of residential images is reduced to 50, including 40 for training, 5 for validation and 5 for test, with other classes unchanged. The result of the experiment is shown in Figure \ref{FigureSydneyfvbalanced}. The generated sentences on balanced Sydney-captions dataset get diversity, which means the imbalance of dataset has a bad influence on the metrics scores of generated sentences. Hence we consider the balance factor when constructing dataset RSICD.

\subsubsection{Generalization capabilities}
\label{analysis:2}
To clarify the generalization capabilities of models trained on different datasets, we illustrate the performance in Table \ref{generalization}. For simplicity, the metrics shown in Table \ref{generalization} are METEOR and CIDEr. UCM\_model represents the model trained on UCM-captions datasets. From Table \ref{generalization}, it can be found that the generalization capabilities of the model trained on RSICD is better than model trained on UCM-captions dataset evaluated on the test images of Sydney-captions dataset. And the generalization capabilities of the model trained on UCM-captions dataset is better than model trained on Sydney-captions dataset evaluated on the test images of RSICD. Compared with the result of model trained on corresponding dataset, the models trained on other dataset suffers a rapid decline in metrics.
\begin{table}[!htb]
\centering
\caption{ Result of models trained on different datasets}
\label{generalization}
\scalebox{1.0}{
\begin{tabular}{|c|c|c|}
\hline
\hline
model & METEOR &  CIDEr\\
\hline
\multicolumn{3}{|c|}{UCM-captions dataset} \\
\hline
UCM\_model & 0.34635 & 2.31314 \\
\hline
Sydney\_model & 0.12727 & 0.16630 \\
\hline
RSICD\_model & 0.10796 & 0.13634 \\
\hline
\multicolumn{3}{|c|}{Sydney-captions dataset} \\
\hline
UCM\_model & 0.09660 & 0.05614 \\
\hline
Sydney\_model & 0.19295 & 0.91644 \\
\hline
RSICD\_model & 0.10381 & 0.08680 \\
\hline
\multicolumn{3}{|c|}{RSICD dataset} \\
\hline
UCM\_model & 0.07416 & 0.08015 \\
\hline
Sydney\_model & 0.08172 & 0.05765 \\
\hline
RSICD\_model & 0.20459 & 1.18011 \\
\hline
\end{tabular}}
\end{table}
From Table \ref{generalization}, the generalization of model trained on Sydney-captions dataset is better than model trained on RSICD dataset on test images of UCM-captions dataset. However, metrics used in Table \ref{generalization} are objective reference methods to evaluate the quality of generated sentences. It is unfair to utilize the objective metrics to judge the generalization capabilities of different model, since the metrics are computed by comparing the generated sentences with the reference sentences in datasets. When computing the objective metrics, the sentences generated by model trained on RSICD and sentences of test images from other datasets are used to calculate. But subjective evaluation criterion can give a far better evaluation of the generated sentences. To evaluate the sentences in subjective criterion, the sentences generated are divided into three levels: ``related to image", ``unrelated to image", and ``totally depict image". ``related to image" means the sentences can capture the main topics of the image, and maybe have some errors; ``unrelated to image" means the sentences have no relation with the main topics of the image; ``totally depict image" means the sentences can depict main topics of the image correctly. The results of models trained on different datasets evaluating on test images of different datasets are shown in Tables \ref{subjectiveevalucm}-\ref{subjectiveevalrsicd}.

\begin{table}[!htb]
\centering
\caption{ Result of subjective evaluation on UCM-captions dataset}
\label{subjectiveevalucm}
\scalebox{1.0}{
\begin{tabular}{|c|c|c|c|}
\hline
\hline
& UCM\_model & Sydney\_model& RSICD\_model\\
\hline
 unrelated to image& 19\% & 86\% & 62\% \\
\hline
unrelated to image  & 20\% & 11\% & 32\% \\
\hline
 totally depict image& 61\% & 3\% & 6\% \\
\hline
\end{tabular}}
\end{table}

\begin{table}[!htb]
\centering
\caption{ Result of subjective evaluation on Sydney-captions dataset}
\label{subjectiveevalsydney}
\scalebox{1.0}{
\begin{tabular}{|c|c|c|c|}
\hline
\hline
& UCM\_model & Sydney\_model& RSICD\_model\\
\hline
 unrelated to image& 79\% & 51\% & 60\% \\
\hline
unrelated to image  & 21\% & 17\% & 40\% \\
\hline
 totally depict image& 0\% & 32\% & 0\% \\
\hline
\end{tabular}}
\end{table}

\begin{table}[!htb]
\centering
\caption{ Result of subjective evaluation on RSICD}
\label{subjectiveevalrsicd}
\scalebox{1.0}{
\begin{tabular}{|c|c|c|c|}
\hline
\hline
& UCM\_model & Sydney\_model& RSICD\_model\\
\hline
 unrelated to image& 84\% & 83\% & 46\%\\
\hline
unrelated to image  & 12\% & 15\% & 28\% \\
\hline
 totally depict image& 4\% & 2\% & 26\% \\
\hline
\end{tabular}}
\end{table}

From Table \ref{subjectiveevalucm}, 38\% sentences generated by model trained on RSICD can describe the main topics of the test images of UCM-captions dataset. From Table \ref{subjectiveevalsydney}, it can be shown that 40\% sentences generated by model trained on RSICD are unrelated to test images of Sydney-captions dataset. So the generalization capability of model trained on RSICD are better than the models trained on other datasets, while almost half of generated sentences are unrelated to the images. The remote sensing image caption generation task needs to be studied in the future work.
\section{Conclusion and future work}
\label{sec:6}
In this paper, we give the instructions to describe remote sensing images comprehensively, and construct a remote sensing image captioning dataset (RSICD). Furthermore, to make the dataset more comprehensive and balanced, we evaluate different caption methods based on handcrafted representations and convolutional features on different datasets. Through extensive experiments, we give benchmarks on our dataset using the BLEU, METEOR, ROUGE\_L and CIDEr metric. The experimental results show that the image caption methods for natural image can be transferred to remote sensing image to obtain only acceptable descriptions. But considering the characteristics of remote sensing images, more works need to be done on remote sensing image caption generation.

In future work, RSICD will be more comprehensive than present version because some of the sentences are obtained by duplicating the existing sentences. And we plan to apply some new techniques in image processing field and natural language processing field to remote sensing image caption generation task.

\ifCLASSOPTIONcaptionsoff
  \newpage
\fi
\bibliographystyle{IEEEtran}
\bibliography{RSCaption}

\begin{thebibliography}{10}
\providecommand{\url}[1]{#1}
\csname url@samestyle\endcsname
\providecommand{\newblock}{\relax}
\providecommand{\bibinfo}[2]{#2}
\providecommand{\BIBentrySTDinterwordspacing}{\spaceskip=0pt\relax}
\providecommand{\BIBentryALTinterwordstretchfactor}{4}
\providecommand{\BIBentryALTinterwordspacing}{\spaceskip=\fontdimen2\font plus
\BIBentryALTinterwordstretchfactor\fontdimen3\font minus
  \fontdimen4\font\relax}
\providecommand{\BIBforeignlanguage}[2]{{%
\expandafter\ifx\csname l@#1\endcsname\relax
\typeout{** WARNING: IEEEtran.bst: No hyphenation pattern has been}%
\typeout{** loaded for the language `#1'. Using the pattern for}%
\typeout{** the default language instead.}%
\else
\language=\csname l@#1\endcsname
\fi
#2}}
\providecommand{\BIBdecl}{\relax}
\BIBdecl

\bibitem{Yanfeng2016Nonlinear}
Y.~Gu, T.~Liu, X.~Jia, J.~A. Benediktsson, and J.~Chanussot, ``Nonlinear
  multiple kernel learning with multi-structure-element extended morphological
  profiles for hyperspectral image classification,'' \emph{IEEE Transaction on
  Geoscience and Remote Sensing}, vol.~54, no.~6, pp. 3235--3247, 2016.

\bibitem{Gu2017Multiple}
Y.~Gu, Q.~Wang, and B.~Xie, ``Multiple kernel sparse representation for
  airborne lidar data classification,'' \emph{IEEE Transactions on Geoscience
  and Remote Sensing}, vol.~PP, no.~99, pp. 1--21, 2017.

\bibitem{Xia2016}
G.~S. Xia, J.~Hu, F.~Hu, B.~Shi, X.~Bai, Y.~Zhong, L.~Zhang, and X.~Lu, ``Aid:
  A benchmark data set for performance evaluation of aerial scene
  classification,'' \emph{IEEE Transactions on Geoscience and Remote Sensing},
  vol.~PP, no.~99, pp. 1--17, 2016.

\bibitem{Yuan2015Substance}
Y.~Yuan, M.~Fu, and X.~Lu, ``Substance dependence constrained sparse nmf for
  hyperspectral unmixing,'' \emph{IEEE Transactions on Geoscience and Remote
  Sensing}, vol.~53, no.~6, pp. 2975--2986, 2015.

\bibitem{Lu2017Remote}
X.~Lu, X.~Zheng, and Y.~Yuan, ``Remote sensing scene classification by
  unsupervised representation learning,'' \emph{IEEE Transactions on Geoscience
  and Remote Sensing}, vol.~PP, no.~99, pp. 1--10, 2017.

\bibitem{lu201503}
Y.~Yuan, X.~Zheng, and X.~Lu, ``Spectral--spatial kernel regularized for
  hyperspectral image denoising,'' \emph{IEEE Transactions on Geoscience and
  Remote Sensing}, vol.~53, no.~7, pp. 3815--3832, 2015.

\bibitem{lu201601}
Z.~Du, X.~Li, and X.~Lu, ``Local structure learning in high resolution remote
  sensing image retrieval,'' \emph{Neurocomputing}, vol. 207, pp. 813--822,
  2016.

\bibitem{lu201701}
M.~Lu, L.~Hu, T.~Yue, Z.~Chen, B.~Chen, X.~Lu, and B.~Xu, ``Penalized linear
  discriminant analysis of hyperspectral imagery for noise removal,''
  \emph{IEEE Geoscience and Remote Sensing Letters}, vol.~14, no.~3, pp.
  359--363, 2017.

\bibitem{lu201702}
G.~Cheng, J.~Han, and X.~Lu, ``Remote sensing image scene classification:
  Benchmark and state of the art,'' \emph{Proceedings of the IEEE}, vol.~PP,
  no.~99, pp. 1--19, 2017.

\bibitem{cheng2016learning}
G.~Cheng, P.~Zhou, and J.~Han, ``Learning rotation-invariant convolutional
  neural networks for object detection in vhr optical remote sensing images,''
  \emph{IEEE Transactions on Geoscience and Remote Sensing}, vol.~54, no.~12,
  pp. 7405--7415, 2016.

\bibitem{yao2016semantic}
X.~Yao, J.~Han, G.~Cheng, X.~Qian, and L.~Guo, ``Semantic annotation of
  high-resolution satellite images via weakly supervised learning,'' \emph{IEEE
  Transactions on Geoscience and Remote Sensing}, vol.~54, no.~6, pp.
  3660--3671, 2016.

\bibitem{han2015object}
J.~Han, D.~Zhang, G.~Cheng, L.~Guo, and J.~Ren, ``Object detection in optical
  remote sensing images based on weakly supervised learning and high-level
  feature learning,'' \emph{IEEE Transactions on Geoscience and Remote
  Sensing}, vol.~53, no.~6, pp. 3325--3337, 2015.

\bibitem{zheng2016target}
X.~Zheng, Y.~Yuan, and X.~Lu, ``A target detection method for hyperspectral
  image based on mixture noise model,'' \emph{Neurocomputing}, vol. 216, pp.
  331--341, 2016.

\bibitem{yuan2017discovering}
Y.~Yuan, X.~Zheng, and X.~Lu, ``Discovering diverse subset for unsupervised
  hyperspectral band selection,'' \emph{IEEE Transactions on Image Processing},
  vol.~26, no.~1, pp. 51--64, 2017.

\bibitem{lu2017joint}
X.~Lu, Y.~Yuan, and X.~Zheng, ``Joint dictionary learning for multispectral
  change detection,'' \emph{IEEE Transactions on cybernetics}, vol.~47, no.~4,
  pp. 884--897, 2017.

\bibitem{cheng2015effective}
G.~Cheng, J.~Han, L.~Guo, Z.~Liu, S.~Bu, and J.~Ren, ``Effective and efficient
  midlevel visual elements-oriented land-use classification using vhr remote
  sensing images,'' \emph{IEEE Transactions on Geoscience and Remote Sensing},
  vol.~53, no.~8, pp. 4238--4249, 2015.

\bibitem{tuia2009active}
D.~Tuia, F.~Ratle, F.~Pacifici, M.~F. Kanevski, and W.~J. Emery, ``Active
  learning methods for remote sensing image classification,'' \emph{IEEE
  Transactions on Geoscience and Remote Sensing}, vol.~47, no.~7, pp.
  2218--2232, 2009.

\bibitem{Yanfeng2015A}
Y.~Gu, Q.~Wang, X.~Jia, and J.~A. Benediktsson, ``A novel mkl model of
  integrating lidar data and msi for urban area classification,'' \emph{IEEE
  Transaction on Geoscience and Remote Sensing}, vol.~53, no.~10, pp.
  5312--5326, 2015.

\bibitem{melgani2004classification}
F.~Melgani and L.~Bruzzone, ``Classification of hyperspectral remote sensing
  images with support vector machines,'' \emph{IEEE Transactions on geoscience
  and remote sensing}, vol.~42, no.~8, pp. 1778--1790, 2004.

\bibitem{penatti2015deep}
O.~A. Penatti, K.~Nogueira, and J.~A. dos Santos, ``Do deep features generalize
  from everyday objects to remote sensing and aerial scenes domains?'' in
  \emph{Proceedings of the IEEE Conference on Computer Vision and Pattern
  Recognition Workshops}, 2015, pp. 44--51.

\bibitem{inglada2007automatic}
J.~Inglada, ``Automatic recognition of man-made objects in high resolution
  optical remote sensing images by svm classification of geometric image
  features,'' \emph{ISPRS Journal of Photogrammetry and Remote Sensing},
  vol.~62, no.~3, pp. 236--248, 2007.

\bibitem{yuan2014remote}
J.~Yuan, D.~Wang, and R.~Li, ``Remote sensing image segmentation by combining
  spectral and texture features,'' \emph{IEEE Transactions on geoscience and
  remote sensing}, vol.~52, no.~1, pp. 16--24, 2014.

\bibitem{meinel2004comparison}
G.~Meinel and M.~Neubert, ``A comparison of segmentation programs for high
  resolution remote sensing data,'' \emph{International Archives of
  Photogrammetry and Remote Sensing}, vol.~35, no. Part B, pp. 1097--1105,
  2004.

\bibitem{fan2009single}
J.~Fan, M.~Han, and J.~Wang, ``Single point iterative weighted fuzzy c-means
  clustering algorithm for remote sensing image segmentation,'' \emph{Pattern
  Recognition}, vol.~42, no.~11, pp. 2527--2540, 2009.

\bibitem{farag2005unified}
A.~A. Farag, R.~M. Mohamed, and A.~El-Baz, ``A unified framework for map
  estimation in remote sensing image segmentation,'' \emph{IEEE Transactions on
  Geoscience and Remote Sensing}, vol.~43, no.~7, pp. 1617--1634, 2005.

\bibitem{Qu2016124}
B.~Qu, X.~Li, D.~Tao, and X.~Lu, ``Deep semantic understanding of high
  resolution remote sensing image,'' \emph{International Conference on
  Computer, Information and Telecommunication Systems}, pp. 124--128, 2016.

\bibitem{lu2017latent}
X.~Lu, X.~Zheng, and X.~Li, ``Latent semantic minimal hashing for image
  retrieval,'' \emph{IEEE Transactions on Image Processing}, vol.~26, no.~1,
  pp. 355--368, 2017.

\bibitem{shi2017can}
Z.~Shi and Z.~Zou, ``Can a machine generate humanlike language descriptions for
  a remote sensing image?'' \emph{IEEE Transactions on Geoscience and Remote
  Sensing}, vol.~PP, no.~99, pp. 1--12, 2017.

\bibitem{Karpathy20153128}
A.~Karpathy and L.~Fei-Fei, ``Deep visual-semantic alignments for generating
  image descriptions,'' \emph{IEEE Conference on Computer Vision and Pattern
  Recognition}, pp. 3128--3137, 2015.

\bibitem{Xu20152048}
K.~Xu, J.~Ba, R.~Kiros, K.~Cho, A.~Courville, R.~Salakhutdinov, R.~Zemel, and
  Y.~Bengio, ``Show, attend and tell: Neural image caption generation with
  visual attention,'' \emph{International Conference on Machine Learning}, pp.
  2048--2057, 2015.

\bibitem{Johnson20164565}
J.~Johnson, A.~Karpathy, and L.~Fei-Fei, ``Densecap: Fully convolutional
  localization networks for dense captioning,'' \emph{IEEE Conference on
  Computer Vision and Pattern Recognition}, pp. 4565--4574, 2016.

\bibitem{ordonez2011im2text}
V.~Ordonez, G.~Kulkarni, and T.~L. Berg, ``Im2text: Describing images using 1
  million captioned photographs,'' in \emph{Advances in Neural Information
  Processing Systems}, 2011, pp. 1143--1151.

\bibitem{Li2012}
S.~Li, G.~Kulkarni, T.~L. Berg, A.~C. Berg, and Y.~Choi, ``Composing simple
  image descriptions using web-scale n-grams,'' in \emph{Fifteenth Conference
  on Computational Natural Language Learning}, 2011, pp. 220--228.

\bibitem{Yang2011444}
Y.~Yang, C.~L. Teo, D.~H. Iii, and Y.~Aloimonos, ``Corpus-guided sentence
  generation of natural images,'' \emph{Conference on Empirical Methods in
  Natural Language Processing}, pp. 444--454, 2011.

\bibitem{Hodosh2013853}
M.~Hodosh, P.~Young, and J.~Hockenmaier, ``Framing image description as a
  ranking task: data, models and evaluation metrics,'' \emph{Journal of
  Artificial Intelligence Research}, vol.~47, no.~1, pp. 853--899, 2013.

\bibitem{Young201467}
P.~Young, A.~Lai, M.~Hodosh, and J.~Hockenmaier, ``From image descriptions to
  visual denotations: New similarity metrics for semantic inference over event
  descriptions,'' \emph{Transactions of the Association for Computational
  Linguistics}, vol.~2, pp. 67--78, 2014.

\bibitem{Chen2015}
X.~Chen, H.~Fang, T.~Y. Lin, R.~Vedantam, S.~Gupta, P.~Dollar, and C.~L.
  Zitnick, ``Microsoft coco captions: Data collection and evaluation server,''
  \emph{Computer Science}, 2015.

\bibitem{Cho2014}
K.~Cho, B.~V. Merrienboer, C.~Gulcehre, D.~Bahdanau, F.~Bougares, H.~Schwenk,
  and Y.~Bengio, ``Learning phrase representations using rnn encoder-decoder
  for statistical machine translation,'' \emph{Proceedings of the Conference on
  Empirical Methods in Natural Language Processing}, pp. 1724--1734, 2014.

\bibitem{Vinyals20153156}
O.~Vinyals, A.~Toshev, S.~Bengio, and D.~Erhan, ``Show and tell: A neural image
  caption generator,'' \emph{IEEE Conference on Computer Vision and Pattern
  Recognition}, pp. 3156--3164, 2015.

\bibitem{Yang20162361}
Z.~Yang, Y.~Yuan, Y.~Wu, R.~Salakhutdinov, and W.~W. Cohen, ``Review networks
  for caption generation,'' \emph{Advances in Neural Information Processing
  Systems}, pp. 2361--2369, 2016.

\bibitem{Yang2010270}
Y.~Yang and S.~Newsam, ``Bag-of-visual-words and spatial extensions for
  land-use classification,'' \emph{ACM SIGSPATIAL International Conference on
  Advances in Geographic Information Systems}, pp. 270--279, 2010.

\bibitem{Zhang20152175}
F.~Zhang, B.~Du, and L.~Zhang, ``Saliency-guided unsupervised feature learning
  for scene classification,'' \emph{IEEE Transactions on Geoscience and Remote
  Sensing}, vol.~53, no.~4, pp. 2175--2184, 2015.

\bibitem{sivic2003video}
J.~Sivic, A.~Zisserman \emph{et~al.}, ``Video google: A text retrieval approach
  to object matching in videos.'' in \emph{IEEE International Conference on
  Computer Vision}, vol.~2, no. 1470, 2003, pp. 1470--1477.

\bibitem{perronnin2010large}
F.~Perronnin, Y.~Liu, J.~Sanchez, and H.~Poirier, ``Large-scale image retrieval
  with compressed fisher vectors,'' in \emph{Computer Vision and Pattern
  Recognition}, 2010, pp. 3384--3391.

\bibitem{Jegou20111704}
H.~Jegou, F.~Perronnin, M.~Douze, J.~Sanchez, P.~Perez, and C.~Schmid,
  ``Aggregating local image descriptors into compact codes,'' \emph{IEEE
  Transactions on Pattern Analysis and Machine Intelligence}, vol.~34, pp.
  1704--1716, 2011.

\bibitem{lowe2004distinctive}
Lowe and D.~G, ``Distinctive image features from scale-invariant keypoints,''
  \emph{International Journal of Computer Vision}, vol.~60, no.~2, pp. 91--110,
  2004.

\bibitem{Krizhevsky20121097}
A.~Krizhevsky, I.~Sutskever, and G.~E. Hinton, ``Imagenet classification with
  deep convolutional neural networks,'' \emph{International Conference on
  Neural Information Processing Systems}, pp. 1097--1105, 2012.

\bibitem{Jia2014675}
Y.~Jia, E.~Shelhamer, J.~Donahue, S.~Karayev, J.~Long, R.~Girshick,
  S.~Guadarrama, and T.~Darrell, ``Caffe: Convolutional architecture for fast
  feature embedding,'' \emph{ACM International Conference on Multimedia}, pp.
  675--678, 2014.

\bibitem{Zeiler2014818}
M.~D. Zeiler and R.~Fergus, ``Visualizing and understanding convolutional
  networks,'' \emph{European Conference on Computer Vision}, pp. 818--833,
  2014.

\bibitem{Simonyan2014}
K.~Simonyan and A.~Zisserman, ``Very deep convolutional networks for
  large-scale image recognition,'' \emph{Computer Science}, 2014.

\bibitem{Szegedy20151}
C.~Szegedy, W.~Liu, Y.~Jia, and P.~Sermanet, ``Going deeper with
  convolutions,'' \emph{IEEE Conference on Computer Vision and Pattern
  Recognition}, pp. 1--9, 2015.

\bibitem{rumelhart1988learning}
D.~E. Rumelhart, G.~E. Hinton, and R.~J. Williams, ``Learning representations
  by back-propagating errors,'' \emph{Cognitive modeling}, vol.~5, no.~3, p.~1,
  1988.

\bibitem{williams1989learning}
R.~J. Williams and D.~Zipser, ``A learning algorithm for continually running
  fully recurrent neural networks,'' \emph{Neural computation}, vol.~1, no.~2,
  pp. 270--280, 1989.

\bibitem{hochreiter1991untersuchungen}
S.~Hochreiter, ``Untersuchungen zu dynamischen neuronalen netzen,'' in
  \emph{Master's Thesis, Institut Fur Informatik, Technische Universitat,
  Munchen}, 1991.

\bibitem{hochreiter1997long}
S.~Hochreiter and J.~Schmidhuber, ``Long short-term memory,'' \emph{Neural
  computation}, vol.~9, no.~8, pp. 1735--1780, 1997.

\bibitem{zeiler2014visualizing}
M.~D. Zeiler and R.~Fergus, ``Visualizing and understanding convolutional
  networks,'' in \emph{European Conference on Computer Vision}.\hskip 1em plus
  0.5em minus 0.4em\relax Springer, 2014, pp. 818--833.

\bibitem{Papineni2002311}
K.~Papineni, S.~Roukos, T.~Ward, and W.~J. Zhu, ``Bleu: A method for automatic
  evaluation of machine translation,'' \emph{Association for Computational
  Linguistics}, pp. 311--318, 2002.

\bibitem{lin2004rouge}
C.~Flick, ``Rouge: A package for automatic evaluation of summaries,'' in
  \emph{The Workshop on Text Summarization Branches Out}, 2004, p.~10.

\bibitem{lavie2014meteor}
M.~Denkowski and A.~Lavie, ``Meteor universal: Language specific translation
  evaluation for any target language,'' in \emph{The Workshop on Statistical
  Machine Translation}, 2014, pp. 376--380.

\bibitem{Vedantam20154566}
R.~Vedantam, C.~L. Zitnick, and D.~Parikh, ``Cider: Consensus-based image
  description evaluation,'' \emph{IEEE Conference on Computer Vision and
  Pattern Recognition}, pp. 4566--4575, 2015.

\bibitem{schmidhuber2002learning}
J.~Schmidhuber, F.~Gers, and D.~Eck, ``Learning nonregular languages: A
  comparison of simple recurrent networks and lstm,'' \emph{Neural
  Computation}, vol.~14, no.~9, pp. 2039--2041, 2002.

\end{thebibliography}

\begin{IEEEbiography}[{\includegraphics[width=1in,height=1.25in,clip,keepaspectratio]{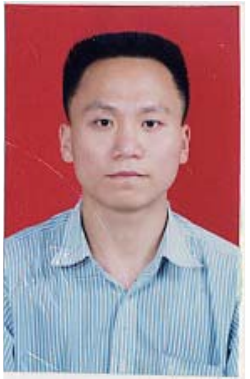}}]{Xiaoqiang Lu} (M'14-SM'15)
is a full professor with the Center for OPTical IMagery Analysis and Learning (OPTIMAL), Xi'an Institute of Optics and Precision Mechanics, Chinese Academy of Sciences, Xi'an, Shaanxi, P. R. China.  His  current research  interests  include  pattern  recognition,  machine  learning,  hyperspectral  image  analysis, cellular automata, and medical imaging.
\end{IEEEbiography}

\begin{IEEEbiography}[{\includegraphics[width=1in,height=1.25in,clip,keepaspectratio]{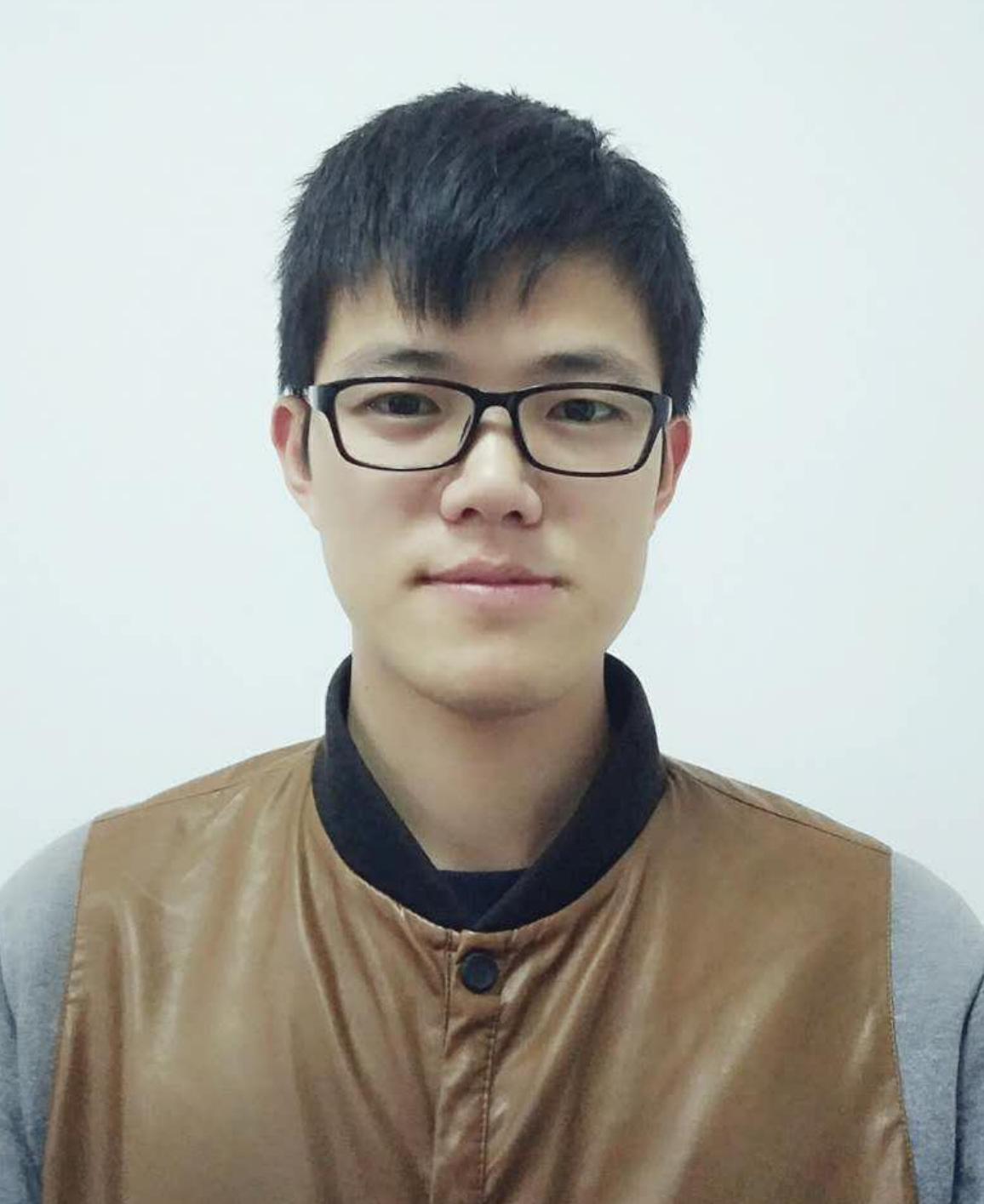}}]{Binqiang Wang}
received the B.S. degree in the school of computer science and engineering, Northwestern Polytechnical University, Xi'an, Shaanxi, P. R. China.
He is currently working toward the Ph.D. degree in signal and information processing in the Center for OPTical IMagery Analysis and Learning (OPTIMAL), Xi'an Institute of Optics and Precision Mechanics, Chinese Academy of Sciences, Xi'an, Shaanxi, P. R. China. His current research interests include pattern recognition, computer vision and machine learning.

\end{IEEEbiography}

\begin{IEEEbiography}[{\includegraphics[width=1in,height=1.25in,clip,keepaspectratio]{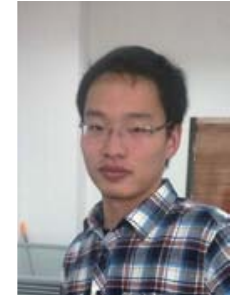}}]{Xiangtao Zheng}
received the M.Sc. and Ph.D. degrees in signal and information processing from the Chinese Academy of Sciences, Xi'an, China, in 2014 and 2017, respectively.

He is currently an Assistant Professor with the Center for OPTical IMagery Analysis and Learning (OPTIMAL), Xi'an Institute of Optics and Precision Mechanics, Chinese Academy of Sciences, Xi'an, Shaanxi, P. R. China. His main research interests include computer vision and pattern recognition.
\end{IEEEbiography}

\begin{IEEEbiographynophoto}
{Xuelong Li}
(M'02-SM'07-F'12) is a full professor with the Xi'an Institute of Optics and Precision Mechanics, Chinese Academy of Sciences, Xi'an 710119, Shaanxi, P. R. China and with The University of Chinese Academy of Sciences, Beijing 100049, P. R. China.
\end{IEEEbiographynophoto}

\end{document}